%% file: main.tex
\documentclass[sigconf]{acmart}

\usepackage{xcolor}
\usepackage{subcaption}
\usepackage{subfiles}
\usepackage{tabularx} %
\usepackage{multirow}
\usepackage{hyperref}
\usepackage{graphicx}
\usepackage{wrapfig}  
\usepackage[utf8]{inputenc} %
\usepackage[T1]{fontenc}    %
\usepackage{hyperref}       %
\usepackage{url}            %
\usepackage{booktabs}       %
\usepackage{amsfonts}       %
\usepackage{nicefrac}       %
\usepackage{microtype}      %
\usepackage{xcolor}         %
\usepackage{enumitem}
\usepackage{makecell}
\usepackage{array}
\usepackage{amsmath, mathrsfs}
\usepackage{amsthm}
\usepackage{cleveref}
\usepackage{color}
\usepackage{comment}
\usepackage[most]{tcolorbox}

\newcommand{\benchmark}{\textsc{HybridRAG-Bench}}
\newcommand{\method}{\textsc{EvoKG}}
\newcommand{\dataset}[1]{\texttt{#1}}

\input{math_commands.tex}

\theoremstyle{plain}

\theoremstyle{definition}

\theoremstyle{remark}

\AtBeginDocument{%
  }

\setcopyright{acmlicensed}
\copyrightyear{2018}
\acmYear{2018}
\acmDOI{XXXXXXX.XXXXXXX}
\acmConference[Conference acronym 'XX]{Make sure to enter the correct
  conference title from your rights confirmation email}{June 03--05,
  2018}{Woodstock, NY}
\acmISBN{978-1-4503-XXXX-X/2018/06}

\begin{document}

\title[A Benchmarking Framework for Multi-Hop Inference over Hybrid Knowledge]{How Much Reasoning Do Retrieval-Augmented Models Add beyond LLMs? A Benchmarking Framework for Multi-Hop Inference over Hybrid Knowledge}

\author{Junhong Lin}
\email{junhong@mit.edu}
\affiliation{%
  \institution{Massachusetts Institute of Technology}
  \city{Cambridge}
  \state{MA}
  \country{United States}
}

\author{Bing Zhang}
\email{Bing.Zhang@ibm.com}
\affiliation{%
  \institution{IBM Research}
  \city{San Jose}
  \state{CA}
  \country{United States}
}

\author{Song Wang}
\email{song.wang@ucf.edu}
\affiliation{%
  \institution{University of Central Florida}
  \city{Orlando}
  \state{FL}
  \country{United States}
}

\author{Ziyan Liu}
\email{zl488@cornell.edu}
\affiliation{%
  \institution{}
  \city{Jersey City}
  \state{NJ}
  \country{United States}
}

\author{Dan Gutfreund}
\email{dgutfre@us.ibm.com}
\affiliation{%
  \institution{IBM Research}
  \city{}
  \state{}
  \country{United States}
}

\author{Julian Shun}
\email{jshun@mit.edu}
\affiliation{%
  \institution{Massachusetts Institute of Technology}
  \city{Cambridge}
  \state{MA}
  \country{United States}
}

\author{Yada Zhu}
\email{yzhu@us.ibm.com}
\affiliation{%
  \institution{IBM Research}
  \city{Yorktown Heights}
  \state{NY}
  \country{United States}
}

\begin{abstract}
Large language models (LLMs) continue to struggle with knowledge-intensive questions that require up-to-date information and multi-hop reasoning. Augmenting LLMs with hybrid external knowledge, such as unstructured text and structured knowledge graphs, offers a promising alternative to costly continual pretraining. %
As such, reliable evaluation of their retrieval and reasoning capabilities becomes critical. However, many existing benchmarks increasingly overlap with LLM pretraining data, which means answers or supporting knowledge may already be encoded in model parameters, making it difficult to distinguish genuine retrieval and reasoning from parametric recall. %
We introduce \benchmark{}, a framework for constructing benchmarks to evaluate retrieval-intensive, multi-hop reasoning over hybrid knowledge. \benchmark{} automatically couples unstructured text and structured knowledge graph representations derived from recent scientific literature on arXiv, and generates knowledge-intensive question-answer pairs grounded in explicit reasoning paths. The framework supports flexible domain and time-frame selection, enabling contamination-aware and customizable evaluation as models and knowledge evolve. 
Experiments across three domains (artificial intelligence, governance and policy, and bioinformatics) %
demonstrate that \benchmark{} rewards genuine retrieval and reasoning rather than parametric recall, offering a principled testbed for evaluating hybrid knowledge–augmented reasoning systems. We release our code and data at \href{https://github.com/junhongmit/HybridRAG-Bench}{github.com/junhongmit/HybridRAG-Bench}.
\end{abstract}


\maketitle

\section{Introduction}
\label{sec:introduction}
\subfile{sections/1_introduction}

\section{Related Works}
\label{sec:related_works}
\subfile{sections/2_related_works}

\section{Problem Definition and Preliminaries}
\label{sec:prelims}
\subfile{sections/3_prelims}

\section{Benchmark Construction}
\label{sec:dataset}
\subfile{sections/4_dataset}

\section{Experiment}
\label{sec:experiment}
\subfile{sections/5_experiment}

\section{Conclusion}
\label{sec:conclusion}
\subfile{sections/6_conclusion}

\bibliographystyle{ACM-Reference-Format}
\bibliography{reference}

\newpage
\appendix
\label{sec:appendix}
\subfile{sections/appendix}

\end{document}

%% file: math_commands.tex
\usepackage{amsmath,amsfonts,bm}

\def\eqref#1{equation~\ref{#1}}

\def\1{\bm{1}}

\DeclareMathAlphabet{\mathsfit}{\encodingdefault}{\sfdefault}{m}{sl}
\SetMathAlphabet{\mathsfit}{bold}{\encodingdefault}{\sfdefault}{bx}{n}

%% file: sections/1_introduction.tex
Large language models (LLMs) increasingly rely on external knowledge to solve knowledge-intensive tasks that require multi-hop reasoning with dispersed evidence. Rather than continually retraining models to absorb newly emerging knowledge, recent systems augment LLMs with external text corpora and structured knowledge graphs (KGs) through retrieval-augmented generation (RAG) and structured knowledge integration (KG-RAG). While these approaches have demonstrated strong empirical gains, a fundamental question remains unresolved: \emph{how much additional \textbf{reasoning} do retrieval-augmented models actually contribute beyond the base LLM?}

Recent studies have indicated that \emph{pretraining contamination}, the overlap between benchmark data and model pretraining corpora, can inflate performance and undermine fair comparison across model generations~\citep{palavalli2024taxonomy, li2024open, dong2024generalization, xu2024benchmark, singh2024evaluation, liang2025much}.
This issue is  severe for hybrid KG-RAG methods, such as Chain-of-Knowledge (CoK)~\citep{lichain}, Reasoning-on-Graph (RoG)~\citep{luoreasoning}, Think-on-Graph (ToG)~\citep{sunthink}, Plan-on-Graph (PoG)~\citep{chenplan}, which are explicitly designed to 
incorporate structured retrieval, graph traversal, and multi-hop reasoning.
To evaluate these methods meaningfully, benchmarks must require models to retrieve relevant evidence from large external sources and compose multiple facts through reasoning; however, the validity of existing benchmarks is increasingly questionable due to both \emph{pretraining contamination} and the difficulty of \emph{continually constructing well-grounded multi-hop questions at scale} to shed light on models' strengths and weaknesses. Many widely-used multi-hop QA benchmarks, such as HotpotQA (2018)~\citep{yang2018hotpotqa},  MetaQA (2018)~\citep{zhang2018variational}, CWQ (2018)~\citep{talmor2018web}, TriviaQA (2017)~\citep{joshi2017triviaqa}, WebQuestions (2013)~\citep{berant2013semantic}, were constructed before the widespread deployment of LLMs and rely on manual curation over static knowledge sources that widely exist in LLM pretraining corpora, making it difficult to isolate the contribution of retrieval and reasoning modules. 

The impact of this overlap is apparent. For example, when asked “\emph{What is the latest film that Denis Villeneuve has been involved in?}”, both Qwen2-72B and Qwen2.5-72B (trained in 2023) incorrectly answer \emph{Dune (2021)}, while Qwen3-32B (trained in 2024) correctly answers \emph{Dune: Part Two (2024)}, despite identical prompting and comparable (even smaller) model parameter size. More broadly, we consider a controlled motivating example by \textbf{directly prompting} them on two CRAG datasets (\dataset{Movie} and \dataset{Sports})~\citep{yang2024crag} developed in March 2024. As shown in \Cref{table:motivation}, the QA accuracy increases sharply (up to +102\%) with later pretraining cutoffs, while improvements on general reasoning benchmarks remain modest (+6.7\%), which cannot solely explain these QA accuracy improvements. When restricting evaluation to questions tagged with involving \textbf{fast-changing facts}, earlier models achieve near-zero accuracy. These results indicate that benchmark performance can be dominated by pretraining exposure rather than retrieval or reasoning ability, obscuring genuine methodological progress. %

\begin{table}[t!]
\caption{Average accuracy (\%) and standard deviation of directly prompting several LLMs with different knowledge cutoff times on the \dataset{Movie} and \dataset{Sports} Q\&A datasets (developed in March 2024), calculated over 5 runs. The best results are highlighted in \textbf{bold}.
}
\label{table:motivation}
\begin{center}
\resizebox{\columnwidth}{!}{%
\newcommand{\res}[2]{#1 \textsubscript{$\pm$ #2}}
\small
\begin{tabular}{cccc}
\toprule
\thead{LLM \\ (Cutoff Time)} &      \thead{Qwen2-72B \\ (June 2023)}   &    \thead{Qwen2.5-72B \\ (October 2023)} & \thead{Qwen3-32B \\ (mid-2024)} \\
\midrule
\dataset{LMArena} \citep{chiang2024chatbot} & 1263 &  1303 (+3.1\%) & \textbf{1347} (+6.7\%) \\
\dataset{OpenLLMLeaderboard} \citep{open-llm-leaderboard-v2} & 43.59\% & 47.98\% (+4.4\%) & -- \\
\midrule
\thead{\normalsize \dataset{Movie} (All)} & \res{24.42}{0.38} &  \res{34.27}{0.32} (+40.3\%) & \textbf{\res{43.86}{1.00}} (+79.6\%) \\
\thead{\normalsize \dataset{Sports} (All)} & \res{16.17}{0.45} & \res{23.78}{0.20} (+47.1\%) & \textbf{\res{32.71}{0.96}} (+102\%) \\
\midrule
\thead{\normalsize \dataset{Movie} (Fast-changing Facts)} & \res{5.38}{2.11} &  \res{15.58}{0.45} (+189\%) & \textbf{\res{22.31}{1.05}} (+315\%) \\
\thead{\normalsize \dataset{Sports} (Fast-changing Facts)} & \res{3.79}{0.34} & \res{11.56}{0.26} (+205\%) & \textbf{\res{27.39}{0.40}} (+622\%) \\
\bottomrule
\end{tabular}
}
\end{center}
\end{table}

\begin{figure*}[t!]
    \centering
    \includegraphics[width=\textwidth]{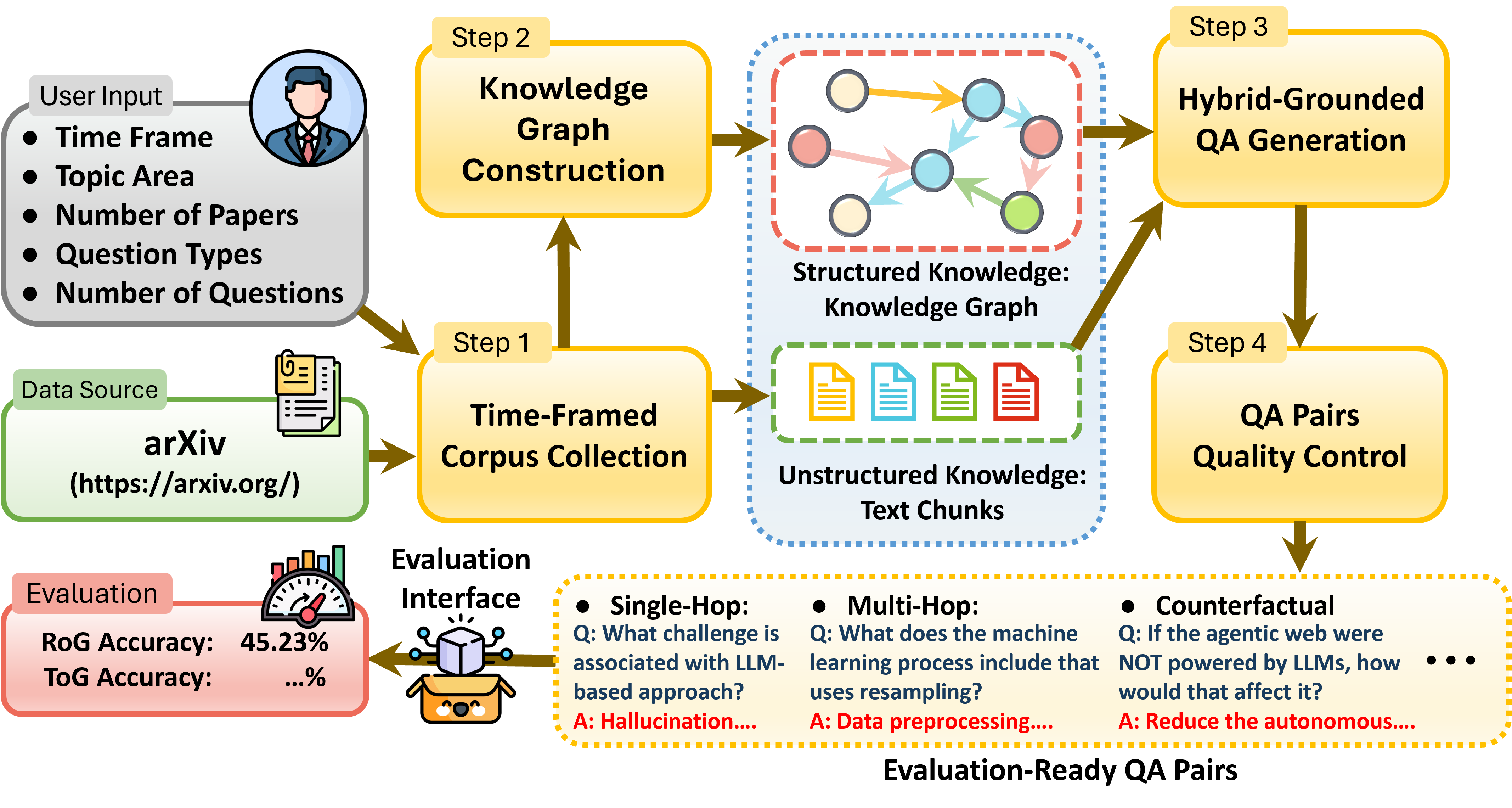}
    \caption{Illustration of the \benchmark{} benchmarking framework.
    }
    \label{fig:framework}
\end{figure*}

To address this gap, we introduce \benchmark{}, a fully automated benchmark construction framework designed to faithfully evaluate retrieval-intensive and multi-hop reasoning. \benchmark{} explicitly externalizes all knowledge required for question answering, thereby satisfying four key properties for valid evaluation. (1) To ensure questions are derived from recent and evolving sources, \benchmark{} collects time-framed scientific corpora from user-specified topics on arXiv, reducing overlap with LLM pretraining data and ensuring that correctness depends on retrieval rather than memorization. (2) To guarantee that questions are knowledge-intensive and sufficiently difficult, the framework constructs a hybrid knowledge environment consisting of aligned text chunks and extracted knowledge graphs, making the location, structure, and provenance of required knowledge explicit. (3) To support diverse reasoning behaviors, \benchmark{} generates challenging question–answer pairs grounded in explicit reasoning paths that cover single-hop lookup, conditional reasoning, multi-hop inference, hard multi-hop chains, counterfactual reasoning, and open-ended synthesis. (4) To maintain scalability, automation, and reusability, \benchmark{} automatically validates QA quality and enables rapid regeneration of benchmarks across new topics, time periods, and knowledge scopes without manual curation. By meeting these criteria, \benchmark{} provides a robust and evolving testbed for evaluating retrieval-augmented and multi-hop reasoning systems.

\benchmark{} makes the following contributions:

\begin{itemize}[leftmargin=15pt, topsep=2pt]
\item \textbf{A benchmarking framework for retrieval-intensive reasoning.}
\benchmark{} is a reusable benchmark construction framework for evaluating multi-hop reasoning over hybrid structured and unstructured knowledge. The framework generates diverse, knowledge-intensive questions grounded in latent knowledge graph paths, yielding challenges that require genuine retrieval and reasoning. The code and data of \benchmark{} is available at \href{https://github.com/junhongmit/HybridRAG-Bench}{github.com/junhongmit/HybridRAG-Bench}.

\item \textbf{Diagnostic benchmark instances for RAG and KG-RAG methods.} 
Using the proposed framework, \benchmark{} instantiates evaluation benchmarks that expose clear and consistent performance gaps across retrieval and reasoning strategies, distinguishing naïve augmentation from effective hybrid reasoning over text and graphs.

\item \textbf{Controlled evaluation of retrieval versus memorization.}
\benchmark{} enables systematic control over domains and document time ranges, allowing principled evaluation of whether performance gains stem from genuine retrieval and multi-hop reasoning rather than parametric memorization.
\end{itemize}

%% file: sections/2_related_works.tex
{\textbf{KGQA Benchmarks.} The Knowledge-Graph Question Answering (KGQA) benchmarks evaluate systems' reasoning and accuracy in handling natural language queries over knowledge graphs. Early benchmarks primarily relied on Freebase ~\citep{bollacker2008freebase} and its subsets. WebQuestions ~\citep{berant2013semantic} provided foundational question-answer pairs, which were re-annotated in WebQSP ~\citep{yih2016value} with semantic parses (SPARQL queries) to allow limiting reasoning to semantically valid paths. CWQ ~\citep{talmor2018web} tested reasoning capabilities on questions of increased compositional complexity. GrailQA ~\citep{gu2021beyond} and KQA Pro ~\citep{cao2022kqa} further expanded question diversity. Because Freebase was discontinued, much of its data has been ingested by modern LLMs during pre-training. Other datasets like MetaQA ~\citep{zhang2018variational}, LC-QuAD ~\citep{trivedi2017lc}, and LC-QuAD 2.0 ~\citep{dubey2019lc} used collaboratively maintained knowledge graphs such as DBpedia ~\citep{lehmann2015dbpedia} and Wikidata ~\citep{vrandevcic2014wikidata}. Recent KG-RAG datasets retrieve and expose graph evidence to LLMs and evaluate the generated natural-language responses. CRAG \citep{yang2024crag} extends beyond Wikipedia to include public data categorized by query dynamism. KGQAGen \citep{zhang2025diagnosing} generates multi-hop QA pairs from local subgraphs and ensures accuracy via SPARQL execution.

As LLMs increasingly incorporate web-scale and scientific corpora during pretraining, KGQA datasets face a growing risk of knowledge overlap with the models' internal memory. Several works have tried to mitigate this issue.
For example, Dynamic-KGQA ~\citep{dammu2025dynamic} attempts to generate adaptive QA pairs using varied reasoning paths within localized sub-knowledge graphs but it does not provide a definitive isolation from parametric memory in LLMs.
Recent works such as ArxivRoll ~\citep{liang2025much} and OKBench ~\citep{li2025okbench} offer dynamic frameworks for generating private evaluation questions from recent documents or news reports; however, it is not designed for KGQA, as it operates on single-document fragments and does not support graph-structured retrieval or multi-fact reasoning.
}

\textbf{Retrieval-Augmented and Knowledge-Grounded LLMs.} To overcome the limitations of parametric knowledge and avoid the cost of continual pretraining, a large body of work has explored augmenting LLMs with external knowledge through retrieval-augmented generation (RAG) or from knowledge graphs. Text-based RAG ~\citep{lewis2020retrieval,asai2024self} methods retrieve unstructured documents to improve factual accuracy, while KG-based approaches leverage structured graphs for relational and multi-hop reasoning. Specifically for knowledge graph reasoning and retrieval: CoK ~\citep{lichain} iteratively retrieves external knowledge from unstructured text, knowledge graphs, and tables into subsequent reasoning steps; RoG ~\citep{luoreasoning} relies on a question-specific subgraph and allows self-correcting erroneous reasoning paths; ToG ~\citep{sunthink} and ToG2.0 ~\citep{mathink} optimize retrieval by performing path expansion and pruning based on seed entities and context; PoG ~\citep{chenplan} and R2-KG ~\citep{jo2025r2} improve answer reliability by enhancing LLM-based judging during exploration and answering. To reason over temporal facts (e.g., TimeQuestions ~\citep{jia2021complex}, MultiTQ ~\citep{chen2023multi}), EvoReasoner ~\citep{lin2025temporal} includes relation and entity reranking based on their temporal alignment with reasoning subgoals.

While these frameworks offer diverse advantages in relational reasoning and hallucination reduction, they are primarily evaluated using static benchmarks such as WebQSP ~\citep{yih2016value}, CWQ ~\citep{talmor2018web}, GrailQA ~\citep{gu2021beyond}, HotpotQA ~\citep{yang2018hotpotqa} and MetaQA ~\citep{zhang2018variational}. These benchmarks are constructed once from Freebase ~\citep{bollacker2008freebase}, DBpedia ~\citep{lehmann2015dbpedia}, and Wikidata ~\citep{vrandevcic2014wikidata}. Consequently, their reported performance can be overestimated due to the overlap between the models' internal parametric knowledge and the external knowledge used for evaluation.

%% file: sections/3_prelims.tex
We formalize the problem addressed by \benchmark{} as \emph{contamination-aware evaluation of knowledge-intensive question answering over hybrid knowledge composed of unstructured text and structured knowledge graphs}. Evaluation is performed within a single domain at a time, while the framework supports instantiation across flexible domains and time ranges.

\subsection{Domains, Corpora, and Knowledge Graphs}

Let $\{\mathcal{D}^{(1)}, \ldots, \mathcal{D}^{(K)}\}$ denote document corpora corresponding to $K$ distinct domains. Each domain $m$ is specified by a set of document selection criteria (e.g., subject categories and keyword constraints), which define the scope of domain-relevant documents.

For each domain $m$, documents are collect within a user-specified time window $[T^{(m)}_{\mathrm{start}}, T^{(m)}_{\mathrm{end}}]$, forming the corpus and its induced evolving knowledge graph:
\[
\mathcal{D}^{(m)} = \{ d^{(m)}_1, \ldots, d^{(m)}_{N_m} \}, \quad
\mathcal{G}^{(m)}_t = (V^{(m)}_t, E^{(m)}_t, \pi^{(m)}_t),
\]
where $\mathcal{G}^{(m)}_t$ represents structured knowledge extracted from documents up to time $t \in [T^{(m)}_{\mathrm{start}}, T^{(m)}_{\mathrm{end}}]$.
Each knowledge graph is constructed independently per domain and may differ in entity types, relation schemas, graph topology, and temporal dynamics. No entities or relations are shared across domains.

\subsection{Questions and Evaluation Scope}

For each domain $m$, we define a set of question--answer pairs
\[
\mathcal{Q}^{(m)} = \{ (q_i, t_i, a_i) \},
\]
where $q_i$ is a question issued at time $t_i$ and $a_i$ is the corresponding ground-truth answer. Each question is evaluated exclusively with respect to the domain-specific knowledge graph snapshot $\mathcal{G}^{(m)}_{t_i}$ and documents available up to time $t_i$.

We assume that all documents, knowledge graph facts, and answers are strictly newer than the pretraining cutoff of the evaluated large language models. Consequently, correct answers cannot be recovered via parametric memorization alone.

\subsection{Task Definition}

Given a question $q \in \mathcal{Q}^{(m)}$ issued at time $t_q$, the task is to predict the answer $a$ using information available up to $t_q$. A model $f$ may reason over:
(1) the knowledge graph snapshot $\mathcal{G}^{(m)}_{t_q}$, and  
(2) the retrieved documents $\mathcal{D}^{(m)}_{t_q} \subseteq \mathcal{D}^{(m)}$. Formally, the model produces
\[
\hat{a} = f\!\left(q, \mathcal{G}^{(m)}_{t_q}, \mathcal{D}^{(m)}_{t_q}\right),
\]
which is evaluated against the ground-truth answer $a$.

%% file: sections/4_dataset.tex
As illustrated in \Cref{fig:framework}, \benchmark{} is a fully automated framework for constructing benchmarks that evaluate retrieval-intensive, multi-hop reasoning over hybrid knowledge. Concretely, \benchmark{} follows a four-stage pipeline.
First, given user-specified topic filters and time ranges, the framework collects time-framed arXiv corpora that define the external knowledge source available to retrieval-based methods.
Second, it constructs hybrid knowledge environments by extracting aligned unstructured text chunks and structured knowledge graphs from the corpus, enabling both document retrieval and graph-based traversal.
Third, the framework generates diverse question-answer pairs grounded in explicit reasoning paths and supporting evidence, covering single-hop, conditional, multi-hop, hard multi-hop, counterfactual, and open-ended reasoning.
Finally, an automated quality control stage filters and validates generated questions to ensure answerability, independence from document-specific phrasing, and non-redundancy.

Together, these components enable reproducible construction of retrieval-intensive benchmarks for evaluating RAG and KG-RAG methods under controlled knowledge settings. We next describe each stage in detail.

\subsection{Time-Framed Corpus Collection}

For each domain, we collect a corpus of arXiv papers based on user-specified document selection criteria, including subject categories (e.g., \texttt{cs.AI}, \texttt{cs.LG}) and optional keyword constraints (e.g., \emph{reinforcement learning}). This mechanism allows \benchmark{} to be instantiated flexibly across domains while maintaining topical coherence.

Document collection is performed over a configurable time window, enabling evaluation under different knowledge cutoff regimes. In our experiments, we select document windows that postdate the public pretraining cutoffs of contemporary LLMs, reducing the likelihood that answers can be recovered via parametric memorization alone.
For each paper, the parsed text, section-level segmentation (e.g., abstract, introduction, methods), and metadata, including title, authors, categories, and submission timestamps, are stored. These documents serve as the unstructured knowledge source for retrieval-based methods.

\subsection{Knowledge Graph Construction}
From the collected corpus, we construct a knowledge graph using \method{}~\citep{lin2025temporal}, a document-driven framework for extracting and organizing structured knowledge from unstructured text. The resulting graph captures entities (e.g., methods, datasets, tasks, policies) and relations expressed in the literature, and is constructed independently for each domain.

\paragraph{Entity Extraction and Alignment.}
Given a batch of documents, \method{} applies a large language model to extract candidate entities and relations. Extracted entities may exhibit lexical variation (e.g., abbreviated method names), semantic ambiguity, or partial descriptions. To address this, \method{} performs context-aware entity alignment, matching each extracted entity against existing KG nodes using joint embeddings over entity type, name, and description. If no existing node exceeds a similarity threshold, a new entity is created; otherwise, the extracted mention is merged with the matched node, preserving provenance information. This alignment step prevents entity fragmentation and ensures consistency across documents and time.

\paragraph{Relation Normalization and Evidence Tracking.}
Extracted relations are normalized to a domain-specific schema and linked to supporting textual evidence. Rather than enforcing a single canonical fact, the knowledge graph retains multiple candidate relations when supported by the corpus, along with confidence scores derived from frequency, recency, and textual support. This design allows the graph to represent uncertainty and variation commonly observed in scientific writing.

The resulting knowledge graph provides structured relational scaffolding that complements the unstructured document corpus, enabling hybrid retrieval and reasoning.

\begin{table}[t!]
\caption{Question statistics by types.}
\label{table:question_types}
\begin{center}
\resizebox{\columnwidth}{!}{%
\newcommand{\res}[2]{#1 \textsubscript{$\pm$ #2}}
\small
\begin{tabular}{cccc}
\toprule
\textbf{Question Type} & \dataset{Arxiv-AI} & \dataset{Arxiv-CY} & \dataset{Arxiv-BIO} \\
\midrule
{Single-hop} & 249 (29\%)  & 238 (25\%) & 264 (25\%) \\
{Single-hop w. Condition} & 139 (16\%) & 128 (13\%) & 193 (19\%)\\
{Multi-hop (Regular)} & 165 (19\%) & 149 (15\%) & 195 (19\%)\\
{Multi-hop (Difficult)} & 149 (17\%) & 166 (17\%) & 139 (13\%)\\
{Counterfactual} & 114 (13\%) & 173 (18\%) & 59 (6\%)\\
{Open-ended} & 47 (5\%) & 112 (12\%) & 190 (18\%)\\
\midrule
{All} & 863 & 966 & 1040\\
\bottomrule
\end{tabular}
}
\end{center}
\end{table}

\subsection{Hybrid-Grounded Question--Answer Generation}
\benchmark{} constructs benchmark questions by grounding them in both \emph{structured reasoning paths from the knowledge graph} and \emph{supporting unstructured textual evidence from the document corpus}. This hybrid grounding ensures that questions cannot be answered solely by graph traversal or document retrieval alone, and instead require effective integration of both modalities.

\paragraph{Reasoning Path Sampling.}
We first sample reasoning paths from the knowledge graph of the form
\[
p = (v_0 \xrightarrow{r_1} v_1 \xrightarrow{r_2} \dots \xrightarrow{r_k} v_k),
\]
where $v_k$ is the answer entity. Each sampled path defines a valid relational structure connecting the question subject to its answer. For every entity and relation along the path, we retrieve the associated supporting text spans from the corpus, which provide contextual descriptions, qualifiers, and evidence.

\paragraph{Hybrid Question Construction.}
For all question types, we condition a large language model on a structured
context that includes:
(1) a sampled reasoning path from the knowledge graph, which specifies the relational constraints,
(2) textual evidence associated with the entities and relations along the path,
and (3) a small set of in-context examples illustrating the desired question
format.
The reasoning path serves as a structural scaffold, while the textual context guides natural language formulation and enables questions that require interpretation or synthesis beyond symbolic lookup.
The model is instructed to generate a question that is faithful to the provided evidence and to produce a ground-truth answer that can be verified from the same context. If the available evidence is insufficient to support a well-posed question (e.g., missing attributes or ambiguous relations), the model is explicitly instructed to abstain from generating an example.

Using this hybrid grounding mechanism, we generate a diverse set of
question--answer pairs. The distribution of question types across domains is
summarized in Table~\ref{table:question_types}.
\begin{itemize}[topsep=1pt, leftmargin=12pt]
\item \textbf{Single-hop questions.} These questions correspond to a single relation lookup $(v_0, r, v_1)$. The generated question directly queries an attribute or relation of the subject entity, while the answer is the terminal entity or value supported by both the KG
edge and its textual evidence.

\item \textbf{Single-hop questions with conditions.} In addition to a single KG relation, these questions include explicit constraints extracted from textual descriptions, such as temporal qualifiers (e.g., ``in 2018''), categorical restrictions, or contextual conditions. The condition is derived from attributes or modifiers mentioned in the supporting text, requiring models to align symbolic relations with unstructured qualifiers.

\item \textbf{Multi-hop questions}. These questions are grounded in reasoning paths with $k \geq 2$ relations. Answering them requires composing multiple relational steps and retrieving evidence across intermediate entities. The question is formulated to obscure the intermediate nodes, requiring implicit multi-hop inference. 

\item \textbf{Difficult multi-hop questions}. To increase difficulty, we preferentially sample paths that traverse high-degree entities in the knowledge graph. Such entities introduce large candidate spaces and increase the likelihood of spurious retrieval, placing greater demands on both retrieval precision and reasoning robustness.

\item \textbf{Counterfactual questions}. Counterfactual questions are constructed by minimally perturbing a relation or attribute in the original reasoning path (e.g., replacing a method, dataset, or condition) and asking how the outcome or conclusion would change. The generated answers are constrained to provide cautious, evidence-based reasoning without introducing unsupported facts.

\item \textbf{Open-ended questions}. Open-ended questions require synthesizing explanations or summaries from multiple pieces of textual evidence aligned with a reasoning path. Rather than returning a single entity, the answer is a short, evidence-grounded natural language response that integrates information from several nodes and relations.
\end{itemize}

\subsection{QA Pairs Quality Control}

Lastly, we apply a multi-stage quality control process to ensure that all generated
question--answer pairs in \benchmark{} are well-posed, answerable, and
non-redundant.

\paragraph{Answerability and Faithfulness.}
Using an LLM-as-a-Judge protocol~\citep{gu2024survey}, we verify that each question can be answered
solely from the provided hybrid context (KG paths and supporting text), and that
the ground-truth answer is faithful to that evidence. Questions requiring
external knowledge or unsupported inference are removed.

\paragraph{Context Independence and Clarity.}
We filter out questions that contain document-local references (e.g., ``in this
paper'', ``Theorem~12'') or are ambiguous or poorly phrased. Retained questions
are lightly normalized (e.g., lowercasing and punctuation standardization)
without altering their semantic content.

Only question--answer pairs that pass all checks are included in the final
benchmark.

%% file: sections/5_experiment.tex
\definecolor{first}{RGB}{237, 106, 89}
\definecolor{second}{RGB}{255, 192, 0}
\definecolor{third}{RGB}{91, 155, 213}
\newcommand{\res}[2]{#1\textsubscript{$\pm$#2}}
\newcommand{\fir}[2]{\textcolor{first}{\textbf{#1\textsubscript{$\pm$#2}}}}
\newcommand{\sed}[2]{\underline{#1\textsubscript{$\pm$#2}}}
\newcommand{\thi}[2]{{#1\textsubscript{$\pm$#2}}}
\newcommand{\firi}[1]{\textcolor{first}{\textbf{#1}}}
\newcommand{\seci}[1]{\underline{#1}}
\newcommand{\thii}[1]{{#1}}
\newcommand{\colorres}[3]{\textcolor{#1}{\textbf{#2} \textsubscript{$\pm$ \textbf{#3}}}}

\begin{table*}[!t]
\caption{Experiment results (accuracy, \%) across LLMs of different scales. We highlight the \firi{first} and \seci{second} best results.}
\label{table:results}
\begin{center}
\renewcommand{\arraystretch}{1.1}
\small
\resizebox{1.0\textwidth}{!}{%
\begin{tabularx}{1.05\textwidth}{@{}c@{\hspace{0.1em}}c@{\hspace{0.2em}}c@{\hspace{0.5em}}c@{\hspace{0.5em}}c@{\hspace{0.5em}}c@{\hspace{0.5em}}c@{\hspace{0.5em}}c@{\hspace{0.5em}}c@{\hspace{0.5em}}c@{\hspace{0.5em}}c@{\hspace{0.5em}}c@{\hspace{0.5em}}c@{\hspace{0.5em}}c@{}}
\toprule
& \thead{\normalsize Datasets $\rightarrow$} & \multicolumn{4}{c}{{\dataset{Arxiv-AI}}} &  \multicolumn{4}{c}{{\dataset{Arxiv-CY}}} &  \multicolumn{4}{c}{{\dataset{Arxiv-BIO}}}\\
\cmidrule(lr){3-6} \cmidrule(lr){7-10} \cmidrule(lr){11-14}
& \thead{\normalsize Methods$\downarrow$}
& \thead{Deep-Seek V3.2\\685B} & \thead{Qwen 2.5\\72B} & \thead{LLaMA 3.3\\70B} & \thead{LLaMA 3.1\\8B} 
& \thead{Deep-Seek V3.2\\685B} & \thead{Qwen 2.5\\72B} & \thead{LLaMA 3.3\\70B} & \thead{LLaMA 3.1\\8B} 
& \thead{Deep-Seek V3.2\\685B} & \thead{Qwen 2.5\\72B} & \thead{LLaMA 3.3\\70B} & \thead{LLaMA 3.1\\8B} 
\\
\midrule
& {IO}
& \res{37.75}{0.64} & \res{27.07}{0.47} & \res{34.11}{0.67} & \res{23.08}{0.64}
& \res{43.62}{0.62} & \res{27.83}{0.20} & \res{35.32}{0.17} & \res{24.06}{0.28}
& \res{39.88}{0.56} & \res{22.81}{0.16} & \res{34.63}{0.08} & \res{25.13}{0.66}
\\
& {CoT}
& \res{36.69}{0.31} & \res{35.32}{0.35} & \res{35.32}{0.58} & \res{27.65}{0.63}
& \res{39.50}{0.75} & \res{39.01}{0.63} & \res{40.64}{0.59} & \res{28.47}{1.04}
& \res{39.75}{0.91} & \res{36.33}{0.08} & \res{37.46}{0.55} & \res{29.96}{0.63}
\\
& {SC}
& \res{35.41}{0.63} & \res{34.02}{0.72} & \res{36.27}{0.88} & \res{26.95}{0.85}
& \res{36.89}{1.49} & \res{37.62}{0.63} & \res{41.24}{0.42} & \res{29.75}{0.70}
& \res{37.29}{0.90} & \res{35.15}{0.64} & \res{37.19}{0.86} & \res{29.35}{0.86}
\\
& {RAG}
& \res{43.68}{0.93} & \res{50.44}{0.07} & \res{47.30}{0.15} & \res{38.32}{0.27}
& \res{41.82}{0.21} & \res{40.99}{4.94} & \res{42.99}{0.23} & \res{35.56}{0.07}
& \res{48.61}{0.75} & \res{52.16}{0.07} & \res{49.18}{0.48} & \res{43.85}{0.14}
\\
\midrule
& {1-hop KG}
& \res{27.55}{0.31} & \res{22.69}{0.19} & \res{26.19}{0.34} & \res{22.57}{0.38}
& \res{29.63}{0.28} & \res{21.86}{0.24} & \res{23.48}{0.20} & \res{23.56}{0.70}
& \res{40.67}{0.66} & \res{25.15}{0.19} & \res{29.75}{0.20} & \res{28.77}{0.28}
\\
& \footnotesize{RAG+1-hop KG}
& \res{49.42}{0.08} & \res{50.64}{0.20} & \res{51.49}{0.18} & \res{43.68}{0.12}
& \res{47.27}{0.52} & \res{47.48}{0.16} & \res{45.69}{1.66} & \res{42.55}{0.29}
& \res{58.99}{0.34} & \res{57.16}{0.07} & \res{54.57}{0.07} & \res{51.20}{0.61}
\\
& {CoK}
& \res{23.08}{0.59} & \res{42.69}{1.33} & \res{38.49}{0.80} & \res{35.62}{2.61}
& \res{27.23}{0.93} & \res{40.43}{0.69} & \res{39.01}{0.74} & \res{36.89}{0.97}
& \res{32.04}{0.89} & \res{41.21}{5.22} & \res{39.90}{0.34} & \res{35.98}{0.67}
\\
& {RoG}
& \res{34.79}{0.38} & \res{37.91}{0.62} & \res{43.55}{0.36} & \res{25.68}{0.48}
& \res{40.81}{0.38} & \res{40.91}{0.14} & \res{44.24}{0.48} & \res{26.15}{1.25}
& \res{36.96}{0.73} & \res{38.67}{0.99} & \res{40.02}{0.16} & \res{24.21}{1.18}
\\
& {ToG}
& \res{40.69}{0.78} & \res{44.50}{0.40} & \res{39.91}{0.23} & \res{17.54}{1.22}
& \res{39.34}{0.45} & \res{37.33}{0.25} & \res{34.58}{0.56} & \res{15.65}{2.64}
& \res{44.15}{0.34} & \res{44.69}{0.29} & \res{42.23}{0.34} & \res{16.06}{5.88}
\\
& {ToG2.0}
& \sed{63.59}{0.80} & \sed{58.61}{0.31} & \sed{59.37}{0.53} & \sed{49.25}{0.51}
& \sed{66.17}{0.47} & \sed{59.05}{0.45} & \sed{60.85}{0.31} & \sed{51.91}{0.84}
& \sed{64.69}{0.95} & \sed{60.36}{0.39} & \sed{57.40}{8.39} & \sed{51.58}{1.83}
\\
& {PoG}
& \res{38.70}{0.27} & \res{43.13}{0.38} & \res{39.86}{0.22} & \res{16.55}{1.05}
& \res{37.37}{0.50} & \res{35.94}{0.40} & \res{34.02}{0.32} & \res{15.78}{2.04}
& \res{42.13}{0.71} & \res{44.10}{0.28} & \res{41.79}{0.34} & \res{21.23}{2.30}
\\
& {R2-KG}
& \res{49.01}{0.99} & \res{42.97}{0.92} & \res{46.51}{0.42} & \res{25.93}{1.55}
& \res{42.67}{0.91} & \res{43.81}{0.70} & \res{46.07}{0.53} & \res{27.76}{1.55}
& \res{50.15}{0.73} & \res{44.25}{3.21} & \res{48.73}{0.59} & \res{27.65}{1.26}
\\
& {HippoRAG2.0}
& \res{62.39}{0.13} & \res{19.47}{2.92} & \res{56.11}{0.47} & \res{47.86}{0.12}
& \res{58.57}{0.09} & \res{12.24}{0.88} & \res{56.00}{0.59} & \res{48.90}{0.21}
& \res{62.19}{0.11} & \res{32.81}{0.16} & \res{56.69}{0.20} & \res{50.33}{0.09}
\\
& {EvoReasoner}
& \fir{75.69}{0.55} & \fir{71.30}{1.36} & \fir{66.78}{0.67} & \fir{55.74}{0.82}
& \fir{69.15}{0.59} & \fir{66.89}{6.13} & \fir{66.31}{0.75} & \fir{55.22}{0.28}
& \fir{75.92}{0.72} & \fir{73.31}{0.55} & \fir{69.94}{0.36} & \fir{58.13}{1.07}
\\
\bottomrule
\end{tabularx}
}
\end{center}
\end{table*}

We design our experiments to evaluate \benchmark{} as a retrieval-intensive benchmark for knowledge-grounded reasoning and to assess whether it meaningfully differentiates between LLM-only, RAG-based, and KG-RAG reasoning approaches. In particular, we seek to answer the following research questions.
\begin{itemize}[topsep=1pt, leftmargin=12pt]
\item \textbf{RQ1:} Are \benchmark{} questions genuinely challenging across LLM scales?
\item \textbf{RQ2:} Does \benchmark{} require external retrieval to answer correctly?
\item \textbf{RQ3:} Does structured knowledge (KG) provide complementary value beyond text-only retrieval?
\item \textbf{RQ4:} Does \benchmark{} discriminate between different retrieval and reasoning strategies?
\end{itemize}

\subsection{Experimental Setup}
\paragraph{Benchmark Instantiation.}
We instantiate \benchmark{} on three domain-specific corpora collected from
recent arXiv papers: AI (reinforcement learning), CY (governance and policy), and
BIO (bioinformatics). For each domain, we construct a domain-specific knowledge
graph and generate approximately 1,000 question--answer pairs grounded in
explicit graph paths and supporting textual evidence. As shown in \Cref{table:question_types}, the questions cover
single-hop, conditional single-hop, multi-hop, hard multi-hop, counterfactual,
and open-ended reasoning. We manually inspect a subset of the generated data to
verify question quality and faithfulness. For answer evaluation, we adopt an LLM-as-a-Judge protocol following CRAG~\citep{yang2024crag}, using the same judge prompt and evaluation models to assess answer correctness and faithfulness.

\textbf{Baselines.} We use four state-of-the-art LLMs with different parameter sizes and knowledge cutoff, including DeepSeek V3.2 (685B, Jul 2024)~\citep{liu2025deepseek}, Qwen 2.5 (72B, Oct 2023)~\citep{qwen2.5}, LLaMA 3.3 (70B, December 2023), and LLaMA 3.1 (8B, Dec 2023)~\citep{grattafiori2024llama}.
We compare a broad range of baselines that differ in how they retrieve and integrate external knowledge: 
(1) LLM-only prompting, without access to external knowledge, including IO-prompt (IO)~\citep{brown2020gpt3}, Chain-of-Thought (CoT)~\citep{wei2022chain}, Self-Consistency (SC)~\citep{wangself};
(2) Text-based RAG, which retrieves relevant passages from the full arXiv corpus using dense retrieval, including Retrieval-Augmented Generation (RAG)~\citep{lewis2020retrieval};
(3) Naive KG augmentation, where local graph neighborhoods (e.g., one-hop neighbors) are directly injected into the prompt, including 1-hop KG~\citep{yang2024crag} (augmenting the LLM with facts from 1-hop KG neighbors of the topic entities) and RAG + 1-hop KG~\citep{yang2024crag};
(4) Hybrid KG-RAG methods, which jointly leverage graph structure and textual evidence through structured retrieval and multi-hop reasoning, including Chain-of-Knowledge (CoK)~\citep{lichain}, Reasoning-on-Graph (RoG)~\citep{luoreasoning}, Think-on-Graph (ToG)~\citep{sunthink}, Think-on-Graph 2.0 (ToG2.0)~\citep{mathink}, Plan-on-Graph (PoG)~\citep{chenplan}, R2-KG~\citep{jo2025r2}, HippoRAG2.0~\citep{gutierrezrag} and EvoReasoner~\citep{lin2025temporal}. We report the average and
standard deviation over 5 runs for all models.

\subsection{Experiment Results}
\textbf{RQ1: Are \benchmark{} Questions Challenging Across LLM Scales?}
As shown in \Cref{table:results}, across all three domains, LLM-only prompting achieves consistently low accuracy across all three domains, ranging from 23–40\%, even when scaling model size from LLaMA-3.1-8B to DeepSeek-V3.2-685B. While larger models yield modest improvements, performance remains far from saturation. This trend indicates that \benchmark{} questions cannot be reliably answered through parametric knowledge alone, even by frontier-scale LLMs. The limited gains from model scaling suggest that success on \benchmark{} does not primarily stem from memorization or general language competence, but instead requires access to external knowledge and non-trivial reasoning. Overall, these results confirm that \benchmark{} poses a substantial and persistent challenge across LLM scales.

\textbf{RQ2: Does \benchmark{} Require External Retrieval?}
Introducing external retrieval leads to large and consistent performance gains. As observed in \Cref{table:results}, for example, text-based RAG improves accuracy by 7–29 absolute points over LLM-only prompting across most model–domain combinations, with the exception of DeepSeek-V3.2 on \dataset{Arxiv-CY}. 
These gains demonstrate that external knowledge is critical for answering \benchmark{} questions. Without retrieval, models fail to obtain the domain-specific and up-to-date information required by the benchmark.
Notably, naïve KG-based augmentation can \emph{degrade} performance relative to LLM-only prompting. Directly injecting one-hop graph neighborhoods without selective retrieval or structured reasoning introduces substantial noise, distracting the model from relevant evidence. This highlights that \benchmark{} does not merely reward access to more information, but specifically requires \emph{effective retrieval} and evidence selection.

\begin{table*}[!t]
\caption{Qwen2.5-72B Model performance (accuracy, \%) grouped by question types. We highlight the \firi{first} and \seci{second} best results. Due to space limitations, we have omitted the standard deviations, but provide them in the Appendix \ref{sec:additional_results}.}
\label{table:type_results}
\begin{center}
\renewcommand{\arraystretch}{1.1}
\definecolor{first}{RGB}{237, 106, 89}
\definecolor{second}{RGB}{255, 192, 0}
\definecolor{third}{RGB}{91, 155, 213}
\renewcommand{\res}[2]{#1}
\renewcommand{\fir}[2]{\textcolor{first}{\textbf{#1}}}
\renewcommand{\sed}[2]{\underline{#1}}
\small
\resizebox{1.0\textwidth}{!}{%
\begin{tabularx}{1\textwidth}{@{}c@{}c@{\hspace{0.1em}}c@{\hspace{0.1em}}c@{\hspace{0.1em}}c@{\hspace{0.1em}}c@{\hspace{0.1em}}c@{\hspace{0.1em}}c@{\hspace{0.1em}}c@{\hspace{0.1em}}c@{\hspace{0.1em}}c@{\hspace{0.1em}}c@{\hspace{0.1em}}c@{\hspace{0.1em}}c@{\hspace{0.1em}}c@{\hspace{0.1em}}c@{\hspace{0.1em}}c@{\hspace{0.1em}}c@{\hspace{0.1em}}c@{\hspace{0.1em}}c@{\hspace{0.1em}}c@{}}
\toprule
& \thead{\normalsize Datasets $\rightarrow$} & \multicolumn{6}{c}{{\dataset{Arxiv-AI}}} &  \multicolumn{6}{c}{{\dataset{Arxiv-CY}}} &  \multicolumn{6}{c}{{\dataset{Arxiv-BIO}}}\\
\cmidrule(lr){3-8} \cmidrule(lr){9-14} \cmidrule(lr){15-20}
& \thead{\normalsize Methods$\downarrow$}
& \rotatebox{75}{\thead{Single-Hop}} & \rotatebox{75}{\thead{Single-Hop \\ w. Conditions}} & \rotatebox{75}{\thead{Multi-Hop}} & \rotatebox{75}{\thead{Multi-Hop \\ (Difficult)}} & \rotatebox{75}{\thead{Counterfactual}} & \rotatebox{75}{\thead{Open-Ended}}
& \rotatebox{75}{\thead{Single-Hop}} & \rotatebox{75}{\thead{Single-Hop \\ w. Conditions}} & \rotatebox{75}{\thead{Multi-Hop}} & \rotatebox{75}{\thead{Multi-Hop \\ (Difficult)}} & \rotatebox{75}{\thead{Counterfactual}} & \rotatebox{75}{\thead{Open-Ended}}
& \rotatebox{75}{\thead{Single-Hop}} & \rotatebox{75}{\thead{Single-Hop \\ w. Conditions}} & \rotatebox{75}{\thead{Multi-Hop}} & \rotatebox{75}{\thead{Multi-Hop \\ (Difficult)}} & \rotatebox{75}{\thead{Counterfactual}} & \rotatebox{75}{\thead{Open-Ended}} \\ 
\midrule
& {IO}
& \res{21.20}{0.47} & \res{20.29}{0.29} & \res{16.36}{0.77} & \res{17.72}{0.33} & \res{68.07}{1.19} & \res{45.96}{1.70} 
& \res{17.96}{0.18} & \res{16.02}{0.39} & \res{15.10}{0.34} & \res{15.36}{0.67} & \res{68.64}{0.25} & \res{34.15}{0.74} 
& \res{19.70}{0.24} & \res{19.38}{0.25} & \res{7.35}{0.25} & \res{16.81}{0.29} & \res{56.27}{1.27} & \res{40.53}{0.00} 
\\
& {CoT}
& \res{25.22}{0.30} & \res{29.35}{0.54} & \res{25.94}{0.59} & \res{23.76}{0.68} & \res{87.02}{1.02} & \res{50.64}{2.48} 
& \res{27.48}{0.73} & \res{27.34}{0.86} & \res{24.83}{0.42} & \res{20.96}{0.70} & \sed{91.91}{1.32} & \res{40.71}{0.91} 
& \res{32.20}{0.41} & \res{35.34}{0.51} & \res{18.06}{0.69} & \res{22.61}{0.54} & \res{83.39}{1.98} & \res{57.26}{0.70} 
\\
& {SC}
& \res{25.46}{1.21} & \res{26.76}{1.67} & \res{24.00}{0.82} & \res{22.68}{1.30} & \res{84.04}{0.66} & \res{50.64}{2.08} 
& \res{26.30}{0.73} & \res{27.19}{1.74} & \res{23.09}{1.51} & \res{21.45}{1.97} & \res{87.05}{3.22} & \res{40.54}{0.91} 
& \res{29.55}{0.63} & \res{33.47}{1.69} & \res{19.29}{1.09} & \res{21.01}{1.30} & \res{83.05}{3.03} & \res{56.42}{0.70} 
\\
& {RAG}
& \res{53.55}{0.19} & \sed{67.63}{0.59} & \res{41.82}{0.49} & \res{29.31}{0.32} & \res{48.54}{0.41} & \sed{85.11}{0.00} 
& \res{51.68}{0.00} & \sed{68.36}{0.39} & \res{35.23}{1.01} & \res{23.49}{0.00} & \res{23.41}{0.29} & \res{72.32}{0.00} 
& \res{51.70}{0.19} & \res{68.91}{0.00} & \res{34.69}{0.00} & \res{22.10}{0.36} & \res{40.68}{0.00} & \res{79.21}{0.26} 
\\
\midrule
& {1-hop KG}
& \res{33.98}{0.20} & \res{23.88}{0.29} & \res{17.58}{0.00} & \res{14.90}{0.27} & \res{8.07}{0.35} & \res{37.45}{1.70} 
& \res{43.03}{0.21} & \res{24.22}{0.00} & \res{12.62}{1.96} & \res{10.36}{0.59} & \res{1.73}{0.37} & \res{34.64}{0.67} 
& \res{35.61}{0.24} & \res{20.52}{0.25} & \res{17.35}{0.56} & \res{17.54}{0.54} & \res{0.00}{0.00} & \res{36.74}{0.21} 
\\
& \footnotesize{RAG + 1-hop KG}
& \res{58.77}{0.68} & \res{65.47}{0.00} & \res{42.22}{1.03} & \res{32.21}{0.55} & \res{37.72}{0.72} & \res{82.98}{0.00} 
& \sed{65.97}{0.00} & \fir{68.75}{0.00} & \sed{38.03}{0.32} & \res{28.51}{0.28} & \res{15.03}{0.00} & \sed{74.70}{0.42} 
& \sed{61.36}{0.38} & \sed{69.17}{0.26} & \res{35.20}{0.51} & \res{37.68}{0.00} & \res{35.59}{0.00} & \sed{82.63}{0.00} 
\\
& {CoK}
& \res{37.19}{1.60} & \res{41.01}{2.13} & \res{33.21}{1.50} & \res{31.81}{2.90} & \res{74.74}{2.68} & \res{66.81}{3.95} 
& \res{34.45}{2.04} & \res{35.00}{3.22} & \res{28.19}{2.68} & \res{23.61}{2.10} & \res{69.83}{1.73} & \res{55.18}{2.61} 
& \res{42.58}{1.69} & \res{43.63}{2.44} & \res{25.51}{2.16} & \res{29.71}{1.30} & \res{71.53}{4.72} & \res{62.00}{1.71} 
\\
& {RoG}
& \res{37.43}{0.53} & \res{32.66}{1.08} & \res{27.64}{0.62} & \res{30.87}{0.42} & \res{71.23}{2.44} & \res{33.62}{0.85} 
& \res{47.98}{0.81} & \res{32.81}{0.99} & \res{25.77}{0.81} & \res{18.19}{0.96} & \res{71.45}{1.07} & \res{41.79}{1.04} 
& \res{38.79}{3.53} & \res{38.03}{0.41} & \res{22.86}{1.27} & \res{29.13}{0.54} & \res{77.97}{3.39} & \res{50.21}{0.86} 
\\
& {ToG}
& \res{53.09}{0.39} & \res{35.11}{0.70} & \res{38.79}{1.38} & \res{39.60}{0.74} & \res{45.09}{1.31} & \res{60.85}{1.70} 
& \res{57.14}{0.88} & \res{40.31}{1.61} & \res{29.53}{1.20} & \res{25.78}{1.50} & \res{13.99}{0.85} & \res{55.00}{0.91} 
& \res{45.83}{0.71} & \res{40.28}{0.56} & \res{32.78}{0.75} & \res{45.29}{0.36} & \res{45.34}{2.20} & \res{58.68}{0.26} 
\\
& {ToG2.0}
& \sed{59.52}{0.30} & \res{54.82}{0.70} & \sed{44.73}{0.45} & \sed{43.89}{1.32} & \fir{90.35}{1.24} & \res{83.40}{1.59} 
& \res{63.61}{0.57} & \res{55.78}{1.27} & \res{29.53}{1.20} & \sed{29.16}{1.64} & \fir{96.07}{0.43} & \res{69.46}{1.04} 
& \res{60.08}{0.66} & \res{58.24}{0.70} & \sed{38.98}{0.83} & \sed{51.88}{0.58} & \fir{93.56}{1.27} & \res{80.84}{0.42} 
\\
& {PoG}
& \res{51.65}{0.20} & \res{36.12}{0.54} & \res{30.55}{1.56} & \res{35.70}{0.78} & \res{53.68}{0.35} & \res{60.85}{1.70} 
& \res{54.62}{0.46} & \res{38.91}{0.58} & \res{28.19}{0.95} & \res{19.40}{0.59} & \res{20.92}{0.85} & \res{50.89}{0.98} 
& \res{46.21}{0.63} & \res{38.13}{0.78} & \res{29.80}{0.41} & \res{42.17}{0.54} & \res{60.00}{1.36} & \res{58.42}{0.74} 
\\
& {R2KG}
& \res{43.13}{0.87} & \res{37.99}{1.06} & \res{36.36}{1.63} & \res{30.34}{0.89} & \res{73.68}{2.60} & \res{45.53}{1.70} 
& \res{50.34}{1.08} & \res{42.50}{1.61} & \res{30.87}{0.42} & \res{26.14}{0.30} & \res{60.92}{1.25} & \res{48.39}{1.43} 
& \res{50.53}{0.51} & \res{39.38}{0.73} & \res{35.31}{0.59} & \res{32.32}{4.31} & \res{64.07}{3.28} & \res{53.37}{1.08} 
\\
& {HippoRAG2.0}
& \res{17.67}{1.81} & \res{29.78}{3.74} & \res{20.48}{3.88} & \res{5.23}{0.27} & \res{23.68}{1.75} & \res{30.21}{7.66} 
& \res{13.45}{0.00} & \res{19.53}{0.00} & \res{8.59}{1.61} & \res{4.58}{1.69} & \res{14.34}{1.39} & \res{14.29}{0.00} 
& \res{30.30}{0.24} & \res{41.45}{0.00} & \res{22.45}{0.00} & \res{11.88}{0.35} & \res{56.27}{0.68} & \res{46.11}{0.26} 
\\
& {EvoReasoner}
& \fir{71.75}{0.76} & \fir{75.54}{0.59} & \fir{65.86}{3.02} & \fir{53.47}{2.70} & \sed{88.30}{4.20} & \fir{90.78}{4.37} 
& \fir{69.75}{3.40} & \res{63.28}{31.65} & \fir{54.50}{1.23} & \fir{49.04}{1.46} & \res{83.12}{1.23} & \fir{82.86}{2.07} 
& \fir{75.15}{0.78} & \fir{77.82}{1.41} & \fir{56.94}{1.95} & \fir{56.09}{2.03} & \sed{91.53}{1.07} & \fir{89.89}{0.61} 
\\
\bottomrule
\end{tabularx}
}
\end{center}
\end{table*}

\textbf{RQ3: Does Structured Knowledge Complement Text-Based Retrieval?}
Hybrid KG-RAG KG-RAG that jointly leverage structured graph information and unstructured text consistently outperform text-only RAG baselines across all domains (\Cref{table:results}).
While text-based RAG is effective for retrieving descriptive or explanatory information, it struggles with questions that require relational reasoning, entity disambiguation, or multi-hop composition. Structured knowledge graphs provide explicit relational scaffolding that complements unstructured retrieval, enabling more reliable traversal and reasoning.
The consistent advantage of hybrid methods over text-only RAG demonstrates that structured knowledge offers additive value beyond unstructured retrieval alone.
These results confirm that \benchmark{} is sensitive to how models integrate text and graph evidence, making it a suitable testbed for evaluating structured reasoning capabilities in KG-RAG systems.

\textbf{RQ4: Does \benchmark{} Discriminate Between Retrieval and Reasoning Strategies?}
For clarity of analysis, we present per-question-type results (\Cref{table:type_results}) using Qwen2.5-72B as a representative strong open-source model; results for other LLMs exhibit consistent trends and are provided in the Appendix \ref{sec:additional_results}.
\benchmark{} reveals substantial and consistent performance gaps among KG-RAG baselines, far exceeding the gains obtained by scaling LLM size alone (\Cref{table:results}). These gaps persist across domains and are especially pronounced when performance is analyzed by question type (\Cref{table:type_results}), indicating that the benchmark meaningfully differentiates retrieval and reasoning strategies. Several clear and interpretable trends emerge in the breakdown by question type:
\begin{itemize}[topsep=1pt, leftmargin=12pt]
\item \textbf{Multi-hop and hard multi-hop questions.}
KG-RAG methods that explicitly model graph structure (e.g., ToG, ToG2.0, EvoReasoner) consistently outperform text-only RAG. This reflects the importance of structured traversal and relational composition when reasoning requires chaining multiple facts.

\item \textbf{Single-hop and conditional questions.}
While these questions require limited multi-step composition, methods that effectively integrate both unstructured text and structured knowledge (e.g., ToG2.0, EvoReasoner, RAG+1-hop KG) consistently outperform text-only RAG and KG-only baselines. This suggests that even seemingly simple questions benefit from hybrid evidence integration, particularly when entity grounding or conditional constraints must be resolved.

\item \textbf{Open-ended questions.}
Methods with strong unstructured text retrieval perform better, highlighting the role of descriptive evidence and synthesis beyond discrete relational paths.

\item \textbf{Counterfactual questions.}
Performance on counterfactual queries correlates strongly with a model’s reasoning capability rather than raw retrieval. Methods with explicit reasoning mechanisms (e.g., CoT, Self-Consistency, ToG2.0, EvoReasoner) achieve substantially higher accuracy, while naïve KG augmentation (1-hop KG) performs near zero. Inspection of model outputs shows that one-hop KG often induces overly cautious responses (e.g., “I don’t know”), as local graph neighborhoods provide insufficient support for hypothetical reasoning. This highlights that counterfactual questions primarily test reasoning and uncertainty handling, rather than factual retrieval alone.
\end{itemize}

These results indicate that \benchmark{} not only measures overall accuracy, but also diagnoses how different retrieval and reasoning mechanisms succeed or fail across distinct reasoning regimes. By disentangling performance across reasoning regimes, \benchmark{} provides fine-grained diagnostic signals for evaluating KG-RAG systems.

\begin{table}[t!]
\captionof{table}{KG construction quality comparison.}
\label{table:kg_construction}
\vspace{-0.75em}
\begin{center}
\centering
\resizebox{\columnwidth}{!}{%
\begin{tabularx}{0.52\textwidth}{c@{\hspace{0.8em}}c@{\hspace{0.8em}}c@{\hspace{0.8em}}c@{\hspace{0.8em}}c}
\toprule
\thead{\normalsize Methods} & OpenIE & GraphRAG & KGGen & EvoKG (Ours)\\
\midrule
\thead{Facts Captured \\ (Average on the 106 corpus)} &
29.36\% & 47.08\% & 66.46\% & \textbf{71.36\%} \\
\bottomrule
\end{tabularx}
}
\end{center}
\vspace{-1.5em}
\end{table}

\subsection{KG Construction Effectiveness}\label{sec:kg_effectiveness}

Since \benchmark{} relies on automatically constructed knowledge graphs, we evaluate whether our KG construction pipeline reliably recovers factual information from source documents. This analysis isolates the quality of KG extraction itself and does not involve downstream QA or reasoning performance.

We measure \emph{fact recovery rate}, defined as the fraction of verifiable facts from the source corpus that are successfully recovered in the constructed KG:
\[
\text{Recovery Rate} = \frac{\text{Number of facts recovered from corpus}}{\text{Number of verifiable facts in corpus}}.
\]

We conduct this evaluation on the MINE benchmark~\citep{mo2025kggen}, which contains 106 document corpora, each annotated with 15 manually verified facts. A fact is considered recovered if it can be matched to at least one triplet in the constructed KG.

\Cref{table:kg_construction} compares our pipeline with OpenIE, GraphRAG, and the state-of-the-art KGGen method. Our approach achieves the highest recovery rate, recovering approximately 71\% of verifiable facts, outperforming KGGen by about 5 absolute points.

These results indicate that the KG construction component of HybridRAG-Bench is effective at extracting factual structure from text and is competitive with prior state-of-the-art methods. Importantly, this suggests that the difficulty of HybridRAG-Bench does not stem from weak or incomplete knowledge graph construction, but rather from the retrieval and reasoning demands imposed by the benchmark.

\subsection{KG Construction Cost and Scalability}
Since \benchmark{} relies on automated KG construction, we briefly assess the efficiency and scalability of the underlying KG evolution pipeline. Empirically, both extraction latency and token usage grow approximately linearly with document length, consistent with the expected complexity dominated by LLM token processing and bounded merge operations. No superlinear cost growth is observed. In practice, KG construction can be parallelized across documents, substantially reducing end-to-end wall-clock time. Detailed latency and token scaling results are provided in Appendix~\ref{sec:kg_efficiency}. The observed near-linear scaling is consistent with the theoretical complexity of the KG construction pipeline, which is dominated by HNSW-based entity and relation alignment with total complexity $O(n \log N + m \log M)$.

\begin{figure}[ht!]
    \centering
    \includegraphics[width=\linewidth]{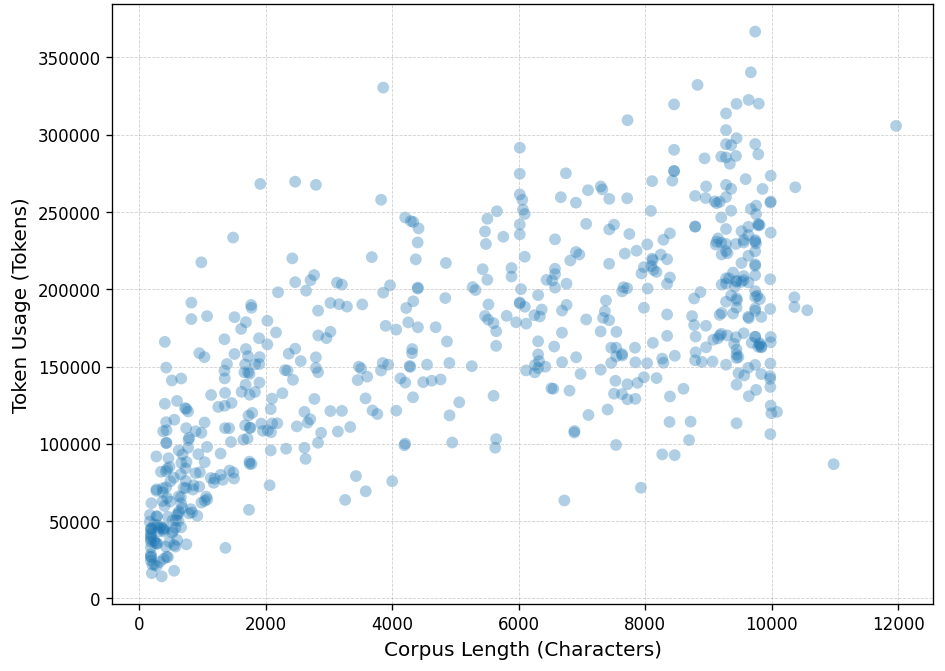}
    \caption{Token usage during KG construction plotted against corpus length. Token cost grows approximately linearly with input size due to proportional input tokens and a fixed number of extraction calls per document. The smooth scaling pattern confirms that EvoKG’s update cost remains predictable and stable across document sizes.}
    \label{fig:token_distribution}
    \vspace{-1em}
\end{figure}

\paragraph{Token Cost Analysis}

\Cref{fig:token_distribution} presents token usage as a function of document length during knowledge graph construction. As shown in the figure, token consumption increases approximately linearly with the size of the input corpus. This behavior arises from two primary factors: (1) input tokens scale proportionally with the raw text length, and (2) output tokens scale smoothly with the semantic density of each document. Because EvoKG performs a fixed and deterministic number of LLM extraction calls per document, the overall token footprint exhibits a consistent slope rather than abrupt jumps or nonlinear spikes. This linear scaling pattern confirms that the computational cost of constructing the KG remains  predictable and bounded. In practice, this  ensures that EvoKG can support long-term or continuously updated corpora without unexpected surges in token usage, making the system practical for real-world deployments where cost control and scalability are critical.

%% file: sections/6_conclusion.tex
We presented \benchmark{}, a benchmark construction framework for evaluating retrieval-intensive reasoning over hybrid knowledge. \benchmark{} is designed to address long-standing challenges in evaluating LLM-based reasoning systems, particularly the difficulty of separating retrieval and reasoning from memorization when benchmarks overlap with model pretraining data.
By constructing hybrid knowledge environments from recent scientific literature and generating questions grounded in explicit reasoning paths and supporting evidence, \benchmark{} enables diagnostic evaluation of LLM-only, RAG-based, and hybrid KG-RAG methods. Our results show that the benchmark poses substantial challenges to LLM-only baselines and meaningfully distinguishes methods based on their ability to retrieve and integrate graph and textual information.
While the framework relies on LLM-based components for knowledge extraction and question generation, and is currently instantiated using scientific literature, we view these choices as practical design decisions that enable scalable, contamination-aware evaluation rather than fundamental limitations. \benchmark{} is intended as reusable research infrastructure, supporting flexible domain and temporal instantiation for future studies of retrieval, reasoning, and generalization over evolving knowledge.

%% file: sections/appendix.tex
\definecolor{first}{RGB}{237, 106, 89}
\definecolor{second}{RGB}{255, 192, 0}
\definecolor{third}{RGB}{91, 155, 213}
\newcommand{\res}[2]{#1\textsubscript{$\pm$#2}}
\newcommand{\fir}[2]{\textcolor{first}{\textbf{#1\textsubscript{$\pm$#2}}}}
\newcommand{\sed}[2]{\underline{#1\textsubscript{$\pm$#2}}}
\newcommand{\thi}[2]{{#1\textsubscript{$\pm$#2}}}
\newcommand{\firi}[1]{\textcolor{first}{\textbf{#1}}}
\newcommand{\seci}[1]{\underline{#1}}
\newcommand{\thii}[1]{{#1}}
\newcommand{\colorres}[3]{\textcolor{#1}{\textbf{#2} \textsubscript{$\pm$ \textbf{#3}}}}

\section{Additional Experiment Results}
\label{sec:additional_results}

\subsection{KG Construction Efficiency and Cost Analysis}
\label{sec:kg_efficiency}
\paragraph{Theoretical Complexity Analysis}
\label{sec:efficiency}
Let $n$ and $m$ denote the number of extracted entities and relations from a corpus, and $N$ and $M$ denote the number of nodes and edges in the existing KG.
KG extraction requires a single LLM call with cost $O(n+m)$.
Entity and relation alignment are performed using HNSW-based similarity search~\citep{malkov2018efficient}, with sublinear complexity $O(n \log N + m \log M)$.
Subsequent merging and insertion operations involve lightweight attribute aggregation and incur linear cost $O(n+m)$.

Overall, the total complexity of KG evolution is $O(n \log N + m \log M)$.
In addition, the pipeline uses a constant number of LLM calls per corpus (five in our implementation), making it practical for continuous and large-scale KG updates.

\begin{figure}[ht!]
    \centering
    \includegraphics[width=\linewidth]{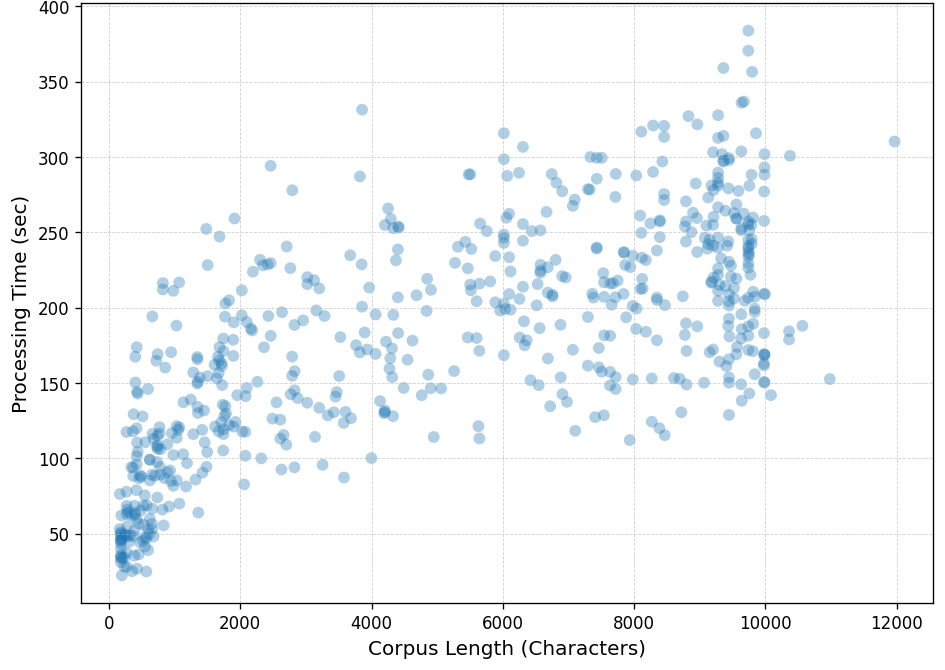}
    \caption{Per-document KG construction latency as a function of corpus length (in characters). Longer documents incur proportionally higher extraction time, and the trend follows the expected near-linear scaling dominated by LLM token processing.}
    \label{fig:time_distribution}
\end{figure}

\paragraph{Empirical Complexity Analysis}
To assess the efficiency of KG construction, we measure the per-document extraction time across all corpora. \Cref{fig:time_distribution} plots the KG-update latency as a function of document length (in characters). The results exhibit a clear positive trend: longer documents incur proportional increases in processing time. This is expected, as extraction latency is dominated by the LLM forward pass, whose computational cost grows with input token count. We observe no evidence of superlinear blow-up. Instead, the empirical trend aligns with the theoretical $O(n \log n + m \log m)$ complexity of EvoKG’s alignment and merge operations. Variance in the vertical direction is attributable to LLM-provider latency: extraction is executed via the OpenRouter API, whose backend routing introduces heterogeneity in response times that is unrelated to algorithmic complexity.

\begin{table*}[t]
\caption{DeepSeek V3.2-685B Model performance (accuracy, \%) grouped by question types. We highlight the \firi{first} and \seci{second} best results.}
\begin{center}
\renewcommand{\arraystretch}{1.1}
\small
\resizebox{1.0\textwidth}{!}{%
\begin{tabularx}{1.32\textwidth}{@{}c@{}c@{\hspace{0.1em}}c@{\hspace{0.1em}}c@{\hspace{0.1em}}c@{\hspace{0.1em}}c@{\hspace{0.1em}}c@{\hspace{0.1em}}c@{\hspace{0.1em}}c@{\hspace{0.1em}}c@{\hspace{0.1em}}c@{\hspace{0.1em}}c@{\hspace{0.1em}}c@{\hspace{0.1em}}c@{\hspace{0.1em}}c@{\hspace{0.1em}}c@{\hspace{0.1em}}c@{\hspace{0.1em}}c@{\hspace{0.1em}}c@{\hspace{0.1em}}c@{\hspace{0.1em}}c@{}}
\toprule
& \thead{\normalsize Datasets $\rightarrow$} & \multicolumn{6}{c}{{\dataset{Arxiv-AI}}} &  \multicolumn{6}{c}{{\dataset{Arxiv-CY}}} &  \multicolumn{6}{c}{{\dataset{Arxiv-BIO}}}\\
\cmidrule(lr){3-8} \cmidrule(lr){9-14} \cmidrule(lr){15-20}
& \thead{\normalsize Methods$\downarrow$}
& \rotatebox{75}{\thead{Single-Hop}} & \rotatebox{75}{\thead{Single-Hop \\ w. Conditions}} & \rotatebox{75}{\thead{Multi-Hop}} & \rotatebox{75}{\thead{Multi-Hop \\ (Difficult)}} & \rotatebox{75}{\thead{Counterfactual}} & \rotatebox{75}{\thead{Open-Ended}}
& \rotatebox{75}{\thead{Single-Hop}} & \rotatebox{75}{\thead{Single-Hop \\ w. Conditions}} & \rotatebox{75}{\thead{Multi-Hop}} & \rotatebox{75}{\thead{Multi-Hop \\ (Difficult)}} & \rotatebox{75}{\thead{Counterfactual}} & \rotatebox{75}{\thead{Open-Ended}}
& \rotatebox{75}{\thead{Single-Hop}} & \rotatebox{75}{\thead{Single-Hop \\ w. Conditions}} & \rotatebox{75}{\thead{Multi-Hop}} & \rotatebox{75}{\thead{Multi-Hop \\ (Difficult)}} & \rotatebox{75}{\thead{Counterfactual}} & \rotatebox{75}{\thead{Open-Ended}} \\ 
\midrule
& {IO}
& \res{30.68}{1.10} & \res{38.85}{1.29} & \res{26.06}{2.30} & \res{23.36}{0.99} & \res{78.25}{2.57} & \res{60.43}{4.58}
& \res{33.70}{0.67} & \res{38.75}{1.27} & \res{30.34}{1.43} & \res{25.18}{0.70} & \sed{85.78}{2.27} & \res{50.18}{1.73}
& \res{33.56}{1.03} & \res{37.72}{1.00} & \res{25.61}{1.09} & \res{26.38}{0.98} & \res{78.31}{2.25} & \res{63.47}{1.18}
\\
& {CoT}
& \res{30.44}{1.23} & \res{34.82}{1.68} & \res{30.42}{1.04} & \res{26.04}{1.43} & \res{69.12}{3.53} & \res{52.34}{2.89}
& \res{29.66}{0.98} & \res{37.03}{3.07} & \res{33.02}{0.99} & \res{25.54}{1.85} & \res{70.29}{1.66} & \res{45.00}{2.56}
& \res{35.61}{1.61} & \res{42.90}{0.83} & \res{25.31}{1.39} & \res{25.51}{1.80} & \res{73.90}{2.75} & \res{56.95}{1.95}
\\
& {SC}
& \res{29.40}{1.20} & \res{35.83}{2.00} & \res{29.70}{2.60} & \res{24.30}{1.55} & \res{66.67}{3.88} & \res{45.53}{2.89}
& \res{28.99}{1.13} & \res{35.94}{2.61} & \res{31.14}{1.83} & \res{21.69}{1.90} & \res{65.20}{1.52} & \res{41.25}{2.49}
& \res{35.91}{1.60} & \res{40.93}{1.70} & \res{19.80}{7.08} & \res{24.78}{2.88} & \res{64.07}{2.49} & \res{51.47}{3.43}
\\
& {RAG}
& \res{48.33}{0.19} & \res{68.11}{0.90} & \res{42.22}{0.57} & \res{27.07}{0.63} & \res{16.37}{3.94} & \res{70.92}{4.37}
& \res{47.69}{0.21} & \res{69.53}{0.00} & \res{41.28}{0.34} & \res{25.30}{0.00} & \res{11.85}{1.45} & \res{70.09}{2.23}
& \res{50.57}{0.19} & \res{69.43}{0.52} & \res{32.91}{1.28} & \res{19.93}{0.36} & \res{15.25}{0.00} & \res{72.11}{1.05}
\\
\midrule
& {1-hop KG}
& \res{37.27}{0.69} & \res{35.83}{0.54} & \res{23.15}{0.59} & \res{18.52}{1.57} & \res{4.39}{0.55} & \res{51.91}{2.55}
& \res{51.76}{0.67} & \res{37.34}{0.58} & \res{21.61}{0.99} & \res{16.02}{0.30} & \res{1.16}{0.37} & \res{48.57}{1.34}
& \res{46.82}{1.03} & \res{43.11}{1.92} & \res{30.20}{1.38} & \res{27.39}{1.33} & \res{2.71}{1.73} & \res{61.89}{0.79}
\\
& \footnotesize{RAG + 1-hop KG}
& \res{59.84}{0.80} & \res{70.86}{0.36} & \res{45.76}{0.30} & \res{35.23}{1.01} & \res{9.21}{0.44} & \res{86.17}{1.06}
& \res{62.82}{1.47} & \res{68.75}{0.00} & \res{43.62}{0.00} & \res{32.83}{0.30} & \res{7.80}{0.29} & \res{74.55}{0.45}
& \sed{64.20}{0.19} & \sed{74.09}{0.00} & \res{43.11}{0.77} & \res{38.04}{0.36} & \res{20.34}{1.69} & \res{80.00}{0.53}
\\
& {CoK}
& \res{24.02}{0.30} & \res{27.77}{1.17} & \res{21.82}{2.17} & \res{15.03}{1.78} & \res{24.39}{2.85} & \res{31.06}{3.18}
& \res{28.99}{1.93} & \res{32.66}{3.14} & \res{24.30}{1.82} & \res{17.47}{3.47} & \res{30.40}{1.13} & \res{30.71}{2.44}
& \res{33.48}{1.74} & \res{39.69}{1.66} & \res{19.80}{1.92} & \res{19.42}{3.19} & \res{29.15}{3.29} & \res{44.95}{1.78}
\\
& {RoG}
& \res{36.55}{1.34} & \res{35.97}{1.02} & \res{28.73}{1.06} & \res{28.99}{1.55} & \res{42.46}{2.05} & \res{42.98}{2.48}
& \res{44.45}{0.62} & \res{42.03}{0.58} & \res{32.62}{1.24} & \res{19.28}{0.93} & \res{59.19}{1.35} & \res{46.07}{0.91}
& \res{40.91}{1.12} & \res{36.68}{1.11} & \res{28.37}{0.83} & \res{26.23}{0.54} & \res{30.17}{4.47} & \res{50.53}{0.88}
\\
& {ToG}
& \res{48.67}{1.00} & \res{47.05}{1.33} & \res{34.30}{2.19} & \res{41.61}{0.95} & \res{22.63}{2.11} & \res{42.98}{1.59}
& \res{56.55}{1.02} & \res{49.84}{1.04} & \res{29.80}{0.91} & \res{35.54}{1.37} & \res{14.10}{0.94} & \res{48.04}{2.96}
& \res{47.54}{1.43} & \res{46.37}{0.45} & \res{35.59}{0.56} & \res{38.59}{1.73} & \res{30.08}{1.41} & \res{54.87}{1.01}
\\
& {ToG2.0}
& \sed{62.01}{1.93} & \res{60.29}{1.90} & \res{47.52}{1.25} & \sed{53.56}{0.99} & \fir{96.32}{1.02} & \sed{90.64}{3.46}
& \sed{69.24}{1.04} & \res{64.69}{0.91} & \sed{44.70}{0.91} & \sed{41.93}{1.77} & \fir{97.11}{0.97} & \res{78.04}{1.45}
& \res{61.44}{0.94} & \res{69.43}{1.70} & \res{44.59}{0.89} & \res{51.88}{2.18} & \fir{95.59}{1.36} & \res{84.84}{0.91}
\\
& {PoG}
& \res{48.84}{0.83} & \res{45.61}{0.98} & \res{33.33}{1.27} & \res{40.40}{1.15} & \res{13.68}{1.31} & \res{38.72}{3.66}
& \res{59.33}{0.81} & \res{46.09}{1.10} & \res{30.20}{1.47} & \res{31.20}{0.70} & \res{5.90}{0.99} & \res{48.04}{1.31}
& \res{47.50}{0.78} & \res{44.04}{0.46} & \res{31.73}{1.59} & \res{40.58}{0.79} & \res{12.20}{3.46} & \res{53.89}{1.58}
\\
& {R2KG}
& \res{47.15}{1.15} & \res{43.88}{1.88} & \res{45.82}{2.56} & \res{45.10}{1.96} & \res{68.42}{1.24} & \res{50.64}{3.66}
& \res{53.61}{0.43} & \res{46.09}{2.96} & \res{36.64}{2.42} & \res{27.47}{2.63} & \res{39.88}{1.42} & \res{50.36}{2.56}
& \res{54.02}{1.72} & \res{49.22}{2.54} & \res{40.71}{1.18} & \res{37.83}{1.06} & \res{68.81}{2.54} & \res{58.63}{1.55}
\\
& {HippoRAG2.0}
& \res{56.39}{0.20} & \sed{73.38}{0.00} & \sed{53.94}{0.38} & \res{38.26}{0.00} & \sed{94.74}{0.00} & \res{89.36}{0.00}
& \res{60.50}{0.00} & \sed{74.22}{0.00} & \res{44.30}{0.00} & \res{32.89}{0.30} & \res{66.01}{0.23} & \sed{82.14}{0.00}
& \res{55.23}{0.15} & \res{65.60}{0.25} & \sed{45.61}{0.25} & \sed{47.83}{0.00} & \sed{94.92}{0.00} & \sed{85.79}{0.00}
\\
& {EvoReasoner}
& \fir{73.80}{0.33} & \fir{83.27}{0.78} & \fir{71.06}{2.20} & \fir{61.74}{1.57} & \res{88.16}{1.32} & \fir{93.62}{2.61}
& \fir{81.37}{1.05} & \fir{79.43}{0.97} & \fir{62.19}{0.84} & \fir{63.05}{1.02} & \res{47.59}{2.88} & \fir{86.31}{0.84}
& \fir{76.52}{1.29} & \fir{81.97}{0.83} & \fir{60.20}{0.85} & \fir{59.57}{2.65} & \res{90.51}{0.83} & \fir{92.53}{1.47}
\\
\bottomrule
\end{tabularx}
}
\end{center}
\end{table*}

\begin{table*}[t]
\caption{Qwen2.5-72B Model performance (accuracy, \%) grouped by question types. We highlight the \firi{first} and \seci{second} best results.}
\begin{center}
\renewcommand{\arraystretch}{1.1}
\small
\resizebox{1.0\textwidth}{!}{%
\begin{tabularx}{1.32\textwidth}{@{}c@{}c@{\hspace{0.1em}}c@{\hspace{0.1em}}c@{\hspace{0.1em}}c@{\hspace{0.1em}}c@{\hspace{0.1em}}c@{\hspace{0.1em}}c@{\hspace{0.1em}}c@{\hspace{0.1em}}c@{\hspace{0.1em}}c@{\hspace{0.1em}}c@{\hspace{0.1em}}c@{\hspace{0.1em}}c@{\hspace{0.1em}}c@{\hspace{0.1em}}c@{\hspace{0.1em}}c@{\hspace{0.1em}}c@{\hspace{0.1em}}c@{\hspace{0.1em}}c@{\hspace{0.1em}}c@{}}
\toprule
& \thead{\normalsize Datasets $\rightarrow$} & \multicolumn{6}{c}{{\dataset{Arxiv-AI}}} &  \multicolumn{6}{c}{{\dataset{Arxiv-CY}}} &  \multicolumn{6}{c}{{\dataset{Arxiv-BIO}}}\\
\cmidrule(lr){3-8} \cmidrule(lr){9-14} \cmidrule(lr){15-20}
& \thead{\normalsize Methods$\downarrow$}
& \rotatebox{75}{\thead{Single-Hop}} & \rotatebox{75}{\thead{Single-Hop \\ w. Conditions}} & \rotatebox{75}{\thead{Multi-Hop}} & \rotatebox{75}{\thead{Multi-Hop \\ (Difficult)}} & \rotatebox{75}{\thead{Counterfactual}} & \rotatebox{75}{\thead{Open-Ended}}
& \rotatebox{75}{\thead{Single-Hop}} & \rotatebox{75}{\thead{Single-Hop \\ w. Conditions}} & \rotatebox{75}{\thead{Multi-Hop}} & \rotatebox{75}{\thead{Multi-Hop \\ (Difficult)}} & \rotatebox{75}{\thead{Counterfactual}} & \rotatebox{75}{\thead{Open-Ended}}
& \rotatebox{75}{\thead{Single-Hop}} & \rotatebox{75}{\thead{Single-Hop \\ w. Conditions}} & \rotatebox{75}{\thead{Multi-Hop}} & \rotatebox{75}{\thead{Multi-Hop \\ (Difficult)}} & \rotatebox{75}{\thead{Counterfactual}} & \rotatebox{75}{\thead{Open-Ended}} \\ 
\midrule
& {IO}
& \res{21.20}{0.47} & \res{20.29}{0.29} & \res{16.36}{0.77} & \res{17.72}{0.33} & \res{68.07}{1.19} & \res{45.96}{1.70} 
& \res{17.96}{0.18} & \res{16.02}{0.39} & \res{15.10}{0.34} & \res{15.36}{0.67} & \res{68.64}{0.25} & \res{34.15}{0.74} 
& \res{19.70}{0.24} & \res{19.38}{0.25} & \res{7.35}{0.25} & \res{16.81}{0.29} & \res{56.27}{1.27} & \res{40.53}{0.00} 
\\
& {CoT}
& \res{25.22}{0.30} & \res{29.35}{0.54} & \res{25.94}{0.59} & \res{23.76}{0.68} & \res{87.02}{1.02} & \res{50.64}{2.48} 
& \res{27.48}{0.73} & \res{27.34}{0.86} & \res{24.83}{0.42} & \res{20.96}{0.70} & \sed{91.91}{1.32} & \res{40.71}{0.91} 
& \res{32.20}{0.41} & \res{35.34}{0.51} & \res{18.06}{0.69} & \res{22.61}{0.54} & \sed{83.39}{1.98} & \res{57.26}{0.70} 
\\
& {SC}
& \res{25.46}{1.21} & \res{26.76}{1.67} & \res{24.00}{0.82} & \res{22.68}{1.30} & \res{84.04}{0.66} & \res{50.64}{2.08} 
& \res{26.30}{0.73} & \res{27.19}{1.74} & \res{23.09}{1.51} & \res{21.45}{1.97} & \res{87.05}{3.22} & \res{40.54}{0.91} 
& \res{29.55}{0.63} & \res{33.47}{1.69} & \res{19.29}{1.09} & \res{21.01}{1.30} & \res{83.05}{3.03} & \res{56.42}{0.70} 
\\
& {RAG}
& \res{53.55}{0.19} & \sed{67.63}{0.59} & \res{41.82}{0.49} & \res{29.31}{0.32} & \res{48.54}{0.41} & \sed{85.11}{0.00} 
& \res{51.68}{0.00} & \sed{68.36}{0.39} & \res{35.23}{1.01} & \res{23.49}{0.00} & \res{23.41}{0.29} & \res{72.32}{0.00} 
& \res{51.70}{0.19} & \res{68.91}{0.00} & \res{34.69}{0.00} & \res{22.10}{0.36} & \res{40.68}{0.00} & \res{79.21}{0.26} 
\\
\midrule
& {1-hop KG}
& \res{33.98}{0.20} & \res{23.88}{0.29} & \res{17.58}{0.00} & \res{14.90}{0.27} & \res{8.07}{0.35} & \res{37.45}{1.70} 
& \res{43.03}{0.21} & \res{24.22}{0.00} & \res{12.62}{1.96} & \res{10.36}{0.59} & \res{1.73}{0.37} & \res{34.64}{0.67} 
& \res{35.61}{0.24} & \res{20.52}{0.25} & \res{17.35}{0.56} & \res{17.54}{0.54} & \res{0.00}{0.00} & \res{36.74}{0.21} 
\\
& \footnotesize{RAG + 1-hop KG}
& \res{58.77}{0.68} & \res{65.47}{0.00} & \res{42.22}{1.03} & \res{32.21}{0.55} & \res{37.72}{0.72} & \res{82.98}{0.00} 
& \sed{65.97}{0.00} & \fir{68.75}{0.00} & \sed{38.03}{0.32} & \res{28.51}{0.28} & \res{15.03}{0.00} & \sed{74.70}{0.42} 
& \sed{61.36}{0.38} & \sed{69.17}{0.26} & \res{35.20}{0.51} & \res{37.68}{0.00} & \res{35.59}{0.00} & \sed{82.63}{0.00} 
\\
& {CoK}
& \res{37.19}{1.60} & \res{41.01}{2.13} & \res{33.21}{1.50} & \res{31.81}{2.90} & \res{74.74}{2.68} & \res{66.81}{3.95} 
& \res{34.45}{2.04} & \res{35.00}{3.22} & \res{28.19}{2.68} & \res{23.61}{2.10} & \res{69.83}{1.73} & \res{55.18}{2.61} 
& \res{42.58}{1.69} & \res{43.63}{2.44} & \res{25.51}{2.16} & \res{29.71}{1.30} & \res{71.53}{4.72} & \res{62.00}{1.71} 
\\
& {RoG}
& \res{37.43}{0.53} & \res{32.66}{1.08} & \res{27.64}{0.62} & \res{30.87}{0.42} & \res{71.23}{2.44} & \res{33.62}{0.85} 
& \res{47.98}{0.81} & \res{32.81}{0.99} & \res{25.77}{0.81} & \res{18.19}{0.96} & \res{71.45}{1.07} & \res{41.79}{1.04} 
& \res{38.79}{3.53} & \res{38.03}{0.41} & \res{22.86}{1.27} & \res{29.13}{0.54} & \res{77.97}{3.39} & \res{50.21}{0.86} 
\\
& {ToG}
& \res{53.09}{0.39} & \res{35.11}{0.70} & \res{38.79}{1.38} & \res{39.60}{0.74} & \res{45.09}{1.31} & \res{60.85}{1.70} 
& \res{57.14}{0.88} & \res{40.31}{1.61} & \res{29.53}{1.20} & \res{25.78}{1.50} & \res{13.99}{0.85} & \res{55.00}{0.91} 
& \res{45.83}{0.71} & \res{40.28}{0.56} & \res{32.78}{0.75} & \res{45.29}{0.36} & \res{45.34}{2.20} & \res{58.68}{0.26} 
\\
& {ToG2.0}
& \sed{59.52}{0.30} & \res{54.82}{0.70} & \sed{44.73}{0.45} & \sed{43.89}{1.32} & \fir{90.35}{1.24} & \res{83.40}{1.59} 
& \res{63.61}{0.57} & \res{55.78}{1.27} & \res{29.53}{1.20} & \sed{29.16}{1.64} & \fir{96.07}{0.43} & \res{69.46}{1.04} 
& \res{60.08}{0.66} & \res{58.24}{0.70} & \sed{38.98}{0.83} & \sed{51.88}{0.58} & \res{93.56}{1.27} & \res{80.84}{0.42} 
\\
& {PoG}
& \res{51.65}{0.20} & \res{36.12}{0.54} & \res{30.55}{1.56} & \res{35.70}{0.78} & \res{53.68}{0.35} & \res{60.85}{1.70} 
& \res{54.62}{0.46} & \res{38.91}{0.58} & \res{28.19}{0.95} & \res{19.40}{0.59} & \res{20.92}{0.85} & \res{50.89}{0.98} 
& \res{46.21}{0.63} & \res{38.13}{0.78} & \res{29.80}{0.41} & \res{42.17}{0.54} & \res{60.00}{1.36} & \res{58.42}{0.74} 
\\
& {R2KG}
& \res{43.13}{0.87} & \res{37.99}{1.06} & \res{36.36}{1.63} & \res{30.34}{0.89} & \res{73.68}{2.60} & \res{45.53}{1.70} 
& \res{50.34}{1.08} & \res{42.50}{1.61} & \res{30.87}{0.42} & \res{26.14}{0.30} & \res{60.92}{1.25} & \res{48.39}{1.43} 
& \res{50.53}{0.51} & \res{39.38}{0.73} & \res{35.31}{0.59} & \res{32.32}{4.31} & \res{64.07}{3.28} & \res{53.37}{1.08} 
\\
& {HippoRAG2.0}
& \res{17.67}{1.81} & \res{29.78}{3.74} & \res{20.48}{3.88} & \res{5.23}{0.27} & \res{23.68}{1.75} & \res{30.21}{7.66} 
& \res{13.45}{0.00} & \res{19.53}{0.00} & \res{8.59}{1.61} & \res{4.58}{1.69} & \res{14.34}{1.39} & \res{14.29}{0.00} 
& \res{30.30}{0.24} & \res{41.45}{0.00} & \res{22.45}{0.00} & \res{11.88}{0.35} & \res{56.27}{0.68} & \res{46.11}{0.26} 
\\
& {EvoReasoner}
& \fir{71.75}{0.76} & \fir{75.54}{0.59} & \fir{65.86}{3.02} & \fir{53.47}{2.70} & \sed{88.30}{4.20} & \fir{90.78}{4.37} 
& \fir{69.75}{3.40} & \res{63.28}{31.65} & \fir{54.50}{1.23} & \fir{49.04}{1.46} & \res{83.12}{1.23} & \fir{82.86}{2.07} 
& \fir{75.15}{0.78} & \fir{77.82}{1.41} & \fir{56.94}{1.95} & \fir{56.09}{2.03} & \fir{91.53}{1.07} & \fir{89.89}{0.61} 
\\
\bottomrule
\end{tabularx}
}
\end{center}
\end{table*}

To clarify practical deployment: although \Cref{fig:time_distribution} reports per-document latency, EvoKG supports data-parallel extraction. When run in parallel, the effective end-to-end KG construction time is substantially lower than the sum of individual extraction times.

Together, \Cref{fig:time_distribution} and \ref{fig:token_distribution} empirically validate the scalability of EvoKG and demonstrate that both latency and token cost grow smoothly and predictably with corpus size.

\subsection{Additional Per-Question-Type Results.}
To provide a more complete view of benchmark behavior across models, we report detailed per-question-type performance for all evaluated LLMs, including DeepSeek-V3.2, Qwen2.5-72B, LLaMA-3.3-70B, and LLaMA-3.1-8B. For each model and baseline, we include accuracy grouped by question type along with standard deviation across 5 runs.

\begin{table*}[!t]
\caption{LLaMa 3.3-70B Model performance (accuracy, \%) grouped by question types. We highlight the \firi{first} and \seci{second} best results.}
\begin{center}
\renewcommand{\arraystretch}{1.1}
\small
\resizebox{1.0\textwidth}{!}{%
\begin{tabularx}{1.32\textwidth}{@{}c@{}c@{\hspace{0.1em}}c@{\hspace{0.1em}}c@{\hspace{0.1em}}c@{\hspace{0.1em}}c@{\hspace{0.1em}}c@{\hspace{0.1em}}c@{\hspace{0.1em}}c@{\hspace{0.1em}}c@{\hspace{0.1em}}c@{\hspace{0.1em}}c@{\hspace{0.1em}}c@{\hspace{0.1em}}c@{\hspace{0.1em}}c@{\hspace{0.1em}}c@{\hspace{0.1em}}c@{\hspace{0.1em}}c@{\hspace{0.1em}}c@{\hspace{0.1em}}c@{\hspace{0.1em}}c@{}}
\toprule
& \thead{\normalsize Datasets $\rightarrow$} & \multicolumn{6}{c}{{\dataset{Arxiv-AI}}} &  \multicolumn{6}{c}{{\dataset{Arxiv-CY}}} &  \multicolumn{6}{c}{{\dataset{Arxiv-BIO}}}\\
\cmidrule(lr){3-8} \cmidrule(lr){9-14} \cmidrule(lr){15-20}
& \thead{\normalsize Methods$\downarrow$}
& \rotatebox{75}{\thead{Single-Hop}} & \rotatebox{75}{\thead{Single-Hop \\ w. Conditions}} & \rotatebox{75}{\thead{Multi-Hop}} & \rotatebox{75}{\thead{Multi-Hop \\ (Difficult)}} & \rotatebox{75}{\thead{Counterfactual}} & \rotatebox{75}{\thead{Open-Ended}}
& \rotatebox{75}{\thead{Single-Hop}} & \rotatebox{75}{\thead{Single-Hop \\ w. Conditions}} & \rotatebox{75}{\thead{Multi-Hop}} & \rotatebox{75}{\thead{Multi-Hop \\ (Difficult)}} & \rotatebox{75}{\thead{Counterfactual}} & \rotatebox{75}{\thead{Open-Ended}}
& \rotatebox{75}{\thead{Single-Hop}} & \rotatebox{75}{\thead{Single-Hop \\ w. Conditions}} & \rotatebox{75}{\thead{Multi-Hop}} & \rotatebox{75}{\thead{Multi-Hop \\ (Difficult)}} & \rotatebox{75}{\thead{Counterfactual}} & \rotatebox{75}{\thead{Open-Ended}} \\ 
\midrule
& {IO}
& \res{31.93}{0.20} & \res{31.83}{0.31} & \res{22.58}{0.50} & \res{20.30}{0.87} & \res{73.03}{1.14} & \res{46.81}{2.13}
& \res{27.39}{0.49} & \res{31.09}{0.91} & \res{29.93}{0.68} & \res{15.90}{0.30} & \res{71.56}{1.18} & \res{36.96}{0.71}
& \res{29.36}{0.33} & \res{32.90}{0.26} & \res{20.28}{0.66} & \res{25.18}{0.60} & \res{77.54}{0.73} & \res{51.97}{1.14}
\\
& {CoT}
& \res{32.73}{0.88} & \res{33.99}{0.60} & \res{25.91}{1.16} & \res{20.30}{0.87} & \res{75.66}{1.14} & \res{39.89}{1.76}
& \res{29.58}{0.43} & \res{33.44}{1.04} & \res{30.60}{1.62} & \res{27.11}{1.01} & \res{80.12}{1.07} & \res{44.82}{0.36}
& \res{34.09}{0.86} & \res{36.68}{0.83} & \res{25.20}{1.10} & \res{28.41}{0.71} & \res{74.24}{1.98} & \res{50.74}{1.55}
\\
& {SC}
& \res{35.14}{1.19} & \res{33.63}{0.93} & \res{27.12}{1.38} & \res{18.96}{0.87} & \res{76.75}{4.04} & \res{39.36}{3.84}
& \res{30.42}{0.57} & \res{33.59}{0.70} & \res{33.56}{0.95} & \res{26.75}{0.90} & \res{79.31}{1.23} & \res{45.89}{2.00}
& \res{33.41}{0.88} & \res{38.45}{1.20} & \res{24.49}{0.97} & \res{26.38}{0.74} & \res{73.56}{1.73} & \res{50.84}{1.51}
\\
& {RAG}
& \res{51.61}{0.45} & \sed{66.73}{0.31} & \res{39.24}{0.50} & \res{18.96}{0.87} & \res{47.37}{0.62} & \res{67.02}{1.84}
& \res{50.28}{0.52} & \res{66.67}{0.37} & \res{37.36}{0.32} & \res{27.31}{0.28} & \res{19.27}{0.72} & \res{67.56}{0.42}
& \res{47.73}{0.76} & \res{65.80}{0.00} & \res{35.71}{0.51} & \res{25.00}{0.36} & \res{44.07}{0.00} & \res{67.37}{0.00}
\\
\midrule
& {1-hop KG}
& \res{32.33}{0.35} & \res{31.47}{0.60} & \res{27.27}{0.43} & \res{21.14}{0.34} & \res{9.65}{0.00} & \res{32.45}{0.92}
& \res{41.76}{0.43} & \res{29.38}{0.62} & \res{23.36}{0.27} & \res{14.34}{0.45} & \res{3.24}{0.28} & \res{22.86}{0.71}
& \res{37.35}{0.39} & \res{31.30}{0.41} & \res{26.63}{0.20} & \res{20.43}{0.54} & \res{8.47}{0.00} & \res{34.21}{0.58}
\\
& \footnotesize{RAG + 1-hop KG}
& \sed{59.64}{0.60} & \res{66.19}{0.00} & \res{46.97}{0.30} & \res{32.21}{0.00} & \res{39.91}{0.44} & \res{71.28}{1.06}
& \res{59.80}{3.07} & \res{63.28}{6.63} & \res{40.94}{2.85} & \res{27.71}{0.98} & \res{14.07}{1.66} & \res{68.15}{3.37}
& \sed{61.55}{0.19} & \sed{67.88}{0.00} & \res{39.29}{0.00} & \res{30.43}{0.00} & \res{43.22}{0.85} & \res{68.16}{0.26}
\\
& {CoK}
& \res{34.94}{1.22} & \res{36.26}{1.33} & \res{29.45}{1.46} & \res{21.14}{0.34} & \res{74.56}{1.66} & \res{48.51}{4.34}
& \res{31.18}{1.17} & \res{32.03}{0.86} & \res{33.15}{2.06} & \res{23.01}{1.39} & \res{72.14}{1.29} & \res{43.93}{1.54}
& \res{34.55}{0.44} & \res{43.73}{1.02} & \res{26.22}{1.78} & \res{24.20}{1.87} & \res{74.58}{4.01} & \res{58.21}{2.17}
\\
& {RoG}
& \res{41.16}{0.45} & \res{39.03}{0.60} & \res{34.24}{0.68} & \res{36.91}{0.47} & \res{76.97}{0.96} & \res{40.43}{1.50}
& \res{49.41}{1.21} & \res{39.06}{0.00} & \res{36.78}{0.50} & \res{24.70}{1.26} & \res{66.59}{0.99} & \res{43.57}{0.67}
& \res{42.95}{0.45} & \res{39.69}{0.70} & \res{29.80}{0.25} & \res{25.80}{0.98} & \res{65.08}{1.73} & \res{49.37}{0.61}
\\
& {ToG}
& \res{50.80}{0.35} & \res{42.27}{1.28} & \res{36.97}{0.74} & \res{42.28}{0.47} & \res{10.53}{1.39} & \res{51.06}{1.50}
& \res{53.28}{0.62} & \res{41.25}{0.31} & \res{32.89}{1.85} & \res{33.13}{1.37} & \res{2.43}{0.43} & \res{41.25}{1.04}
& \res{50.38}{0.76} & \res{44.56}{0.73} & \res{35.92}{0.83} & \res{34.20}{1.16} & \res{11.19}{1.36} & \res{50.53}{1.25}
\\
& {ToG2.0}
& \res{59.44}{0.40} & \res{55.22}{0.93} & \res{46.67}{0.96} & \sed{43.46}{1.20} & \fir{92.54}{2.01} & \fir{85.11}{1.50}
& \sed{64.37}{0.90} & \res{62.03}{0.62} & \res{37.85}{1.73} & \res{34.22}{0.45} & \fir{93.29}{0.69} & \sed{71.96}{1.66}
& \res{35.91}{5.48} & \res{42.38}{3.48} & \res{41.02}{1.10} & \sed{48.99}{1.08} & \fir{91.53}{1.86} & \sed{81.47}{0.52}
\\
& {PoG}
& \res{49.70}{0.66} & \res{41.55}{0.31} & \res{39.39}{0.74} & \res{41.78}{0.56} & \res{11.18}{0.73} & \res{47.87}{1.84}
& \res{53.28}{0.56} & \res{40.31}{0.38} & \res{32.48}{1.24} & \res{31.81}{0.59} & \res{2.54}{0.28} & \res{39.82}{0.91}
& \res{49.55}{0.15} & \res{44.46}{0.51} & \res{34.18}{0.32} & \res{32.75}{1.41} & \res{12.88}{0.83} & \res{51.68}{0.39}
\\
& {R2KG}
& \res{48.90}{0.77} & \res{46.04}{0.51} & \res{38.94}{0.50} & \res{40.44}{1.45} & \res{63.38}{2.73} & \res{39.89}{1.76}
& \res{48.74}{1.16} & \res{51.56}{1.91} & \sed{42.82}{2.18} & \sed{36.02}{0.80} & \res{48.55}{0.73} & \res{49.46}{1.84}
& \res{49.77}{0.66} & \res{52.64}{1.45} & \res{38.78}{1.12} & \res{32.03}{0.54} & \res{70.51}{2.30} & \res{58.95}{1.60}
\\
& {HippoRAG2.0}
& \res{52.01}{0.67} & \res{66.01}{0.93} & \sed{48.33}{0.26} & \res{30.87}{0.00} & \res{87.72}{0.00} & \sed{82.45}{1.76}
& \res{47.14}{0.62} & \sed{71.09}{3.16} & \res{40.40}{0.78} & \res{27.83}{0.96} & \sed{87.28}{0.63} & \res{71.79}{0.91}
& \res{50.98}{0.19} & \res{64.77}{0.00} & \sed{41.12}{0.25} & \res{33.77}{0.58} & \sed{85.42}{1.36} & \res{80.21}{0.26}
\\
& {EvoReasoner}
& \fir{71.29}{0.20} & \fir{76.98}{0.72} & \fir{63.64}{1.21} & \fir{53.36}{1.01} & \res{56.14}{0.88} & \fir{85.11}{0.00}
& \fir{75.29}{1.60} & \fir{77.81}{2.07} & \fir{57.05}{1.53} & \fir{51.08}{1.76} & \res{56.76}{1.43} & \fir{83.75}{1.54}
& \fir{75.57}{1.43} & \fir{76.94}{0.45} & \fir{53.95}{0.91} & \fir{53.44}{0.94} & \res{53.81}{1.85} & \fir{89.34}{1.31}
\\
\bottomrule
\end{tabularx}
}
\end{center}
\end{table*}

\begin{table*}[!t]
\caption{LLaMa 3.1-8B Model performance (accuracy, \%) grouped by question types. We highlight the \firi{first} and \seci{second} best results.}
\begin{center}
\renewcommand{\arraystretch}{1.1}
\small
\resizebox{1.0\textwidth}{!}{%
\begin{tabularx}{1.32\textwidth}{@{}c@{}c@{\hspace{0.1em}}c@{\hspace{0.1em}}c@{\hspace{0.1em}}c@{\hspace{0.1em}}c@{\hspace{0.1em}}c@{\hspace{0.1em}}c@{\hspace{0.1em}}c@{\hspace{0.1em}}c@{\hspace{0.1em}}c@{\hspace{0.1em}}c@{\hspace{0.1em}}c@{\hspace{0.1em}}c@{\hspace{0.1em}}c@{\hspace{0.1em}}c@{\hspace{0.1em}}c@{\hspace{0.1em}}c@{\hspace{0.1em}}c@{\hspace{0.1em}}c@{\hspace{0.1em}}c@{}}
\toprule
& \thead{\normalsize Datasets $\rightarrow$} & \multicolumn{6}{c}{{\dataset{Arxiv-AI}}} &  \multicolumn{6}{c}{{\dataset{Arxiv-CY}}} &  \multicolumn{6}{c}{{\dataset{Arxiv-BIO}}}\\
\cmidrule(lr){3-8} \cmidrule(lr){9-14} \cmidrule(lr){15-20}
& \thead{\normalsize Methods$\downarrow$}
& \rotatebox{75}{\thead{Single-Hop}} & \rotatebox{75}{\thead{Single-Hop \\ w. Conditions}} & \rotatebox{75}{\thead{Multi-Hop}} & \rotatebox{75}{\thead{Multi-Hop \\ (Difficult)}} & \rotatebox{75}{\thead{Counterfactual}} & \rotatebox{75}{\thead{Open-Ended}}
& \rotatebox{75}{\thead{Single-Hop}} & \rotatebox{75}{\thead{Single-Hop \\ w. Conditions}} & \rotatebox{75}{\thead{Multi-Hop}} & \rotatebox{75}{\thead{Multi-Hop \\ (Difficult)}} & \rotatebox{75}{\thead{Counterfactual}} & \rotatebox{75}{\thead{Open-Ended}}
& \rotatebox{75}{\thead{Single-Hop}} & \rotatebox{75}{\thead{Single-Hop \\ w. Conditions}} & \rotatebox{75}{\thead{Multi-Hop}} & \rotatebox{75}{\thead{Multi-Hop \\ (Difficult)}} & \rotatebox{75}{\thead{Counterfactual}} & \rotatebox{75}{\thead{Open-Ended}} \\ 
\midrule
& {IO}
& \res{20.48}{0.92} & \res{19.57}{0.54} & \res{12.48}{1.12} & \res{14.90}{1.49} & \res{56.67}{1.53} & \res{26.38}{2.17}
& \res{16.89}{0.49} & \res{16.25}{0.58} & \res{16.24}{0.27} & \res{12.05}{0.85} & \res{54.22}{1.81} & \res{29.82}{1.07}
& \res{19.85}{0.62} & \res{26.11}{0.96} & \res{11.73}{0.72} & \res{17.39}{0.46} & \res{57.29}{1.66} & \res{40.95}{1.50}
\\
& {CoT}
& \res{25.54}{0.32} & \res{27.34}{1.93} & \res{20.85}{0.73} & \res{17.32}{0.99} & \res{54.56}{1.95} & \res{31.06}{3.18}
& \res{25.71}{1.17} & \res{21.25}{1.51} & \res{24.03}{1.30} & \res{20.36}{1.34} & \res{50.06}{2.98} & \res{27.14}{0.44}
& \res{26.36}{0.98} & \res{31.19}{3.76} & \res{17.86}{0.91} & \res{21.16}{1.80} & \res{51.53}{2.03} & \res{44.21}{2.47}
\\
& {SC}
& \res{26.59}{2.15} & \res{25.76}{0.29} & \res{19.76}{1.94} & \res{16.24}{1.23} & \res{53.86}{4.10} & \res{26.38}{1.70}
& \res{25.38}{1.24} & \res{24.06}{1.88} & \res{26.31}{1.77} & \res{20.24}{1.50} & \res{54.57}{2.27} & \res{25.89}{1.49}
& \res{25.68}{1.98} & \res{31.71}{0.39} & \res{18.57}{1.35} & \res{22.17}{3.06} & \res{57.97}{2.92} & \res{39.47}{2.35}
\\
& {RAG}
& \res{47.59}{0.20} & \res{60.07}{1.80} & \res{34.85}{0.30} & \res{23.15}{0.34} & \res{3.51}{0.00} & \res{68.09}{0.00}
& \res{47.69}{0.21} & \sed{63.28}{0.00} & \res{31.88}{1.01} & \res{22.29}{0.00} & \res{2.31}{0.00} & \res{54.02}{0.45}
& \res{47.35}{0.00} & \res{64.77}{0.00} & \res{30.10}{0.51} & \res{25.00}{0.36} & \res{1.69}{0.00} & \res{58.68}{0.79}
\\
\midrule
& {1-hop KG}
& \res{32.29}{0.32} & \res{25.32}{0.54} & \res{20.12}{0.89} & \res{18.52}{1.73} & \res{0.88}{0.00} & \res{37.02}{1.70}
& \res{45.97}{0.68} & \res{26.56}{1.10} & \res{16.91}{1.15} & \res{11.08}{0.61} & \res{0.00}{0.00} & \res{36.25}{1.66}
& \res{35.61}{0.34} & \res{28.08}{0.21} & \res{20.71}{0.83} & \res{17.10}{1.69} & \res{0.00}{0.00} & \res{45.68}{0.84}
\\
& \footnotesize{RAG + 1-hop KG}
& \fir{53.82}{0.40} & \fir{68.35}{0.72} & \sed{40.91}{0.30} & \res{29.53}{0.67} & \res{3.51}{0.00} & \sed{70.21}{0.00}
& \fir{59.87}{0.21} & \fir{69.53}{0.78} & \res{33.56}{0.00} & \res{28.92}{0.60} & \res{1.73}{0.00} & \sed{70.09}{0.45}
& \fir{57.77}{0.57} & \fir{68.65}{0.26} & \res{32.91}{0.77} & \res{34.06}{0.72} & \res{5.08}{1.69} & \res{70.00}{0.53}
\\
& {CoK}
& \res{29.64}{5.72} & \res{33.67}{2.96} & \res{25.09}{3.03} & \res{23.22}{2.19} & \res{76.84}{3.95} & \res{49.36}{4.34}
& \res{30.25}{2.33} & \res{31.56}{2.60} & \res{25.50}{1.12} & \res{15.90}{1.73} & \sed{76.18}{1.57} & \res{42.68}{3.72}
& \res{34.09}{2.57} & \res{35.75}{2.60} & \res{21.02}{2.24} & \res{23.91}{2.25} & \res{67.12}{4.87} & \res{53.37}{1.40}
\\
& {RoG}
& \res{28.59}{1.06} & \res{22.45}{0.54} & \res{16.12}{1.41} & \res{22.42}{1.83} & \res{42.28}{1.70} & \res{23.40}{1.35}
& \res{32.77}{1.68} & \res{24.22}{1.48} & \res{17.18}{2.19} & \res{10.24}{1.48} & \res{41.73}{1.69} & \res{25.71}{2.96}
& \res{28.64}{1.65} & \res{24.46}{1.52} & \res{13.78}{1.02} & \res{8.12}{1.06} & \res{39.32}{2.92} & \res{35.58}{1.65}
\\
& {ToG}
& \res{19.44}{1.21} & \res{23.74}{2.03} & \res{13.09}{1.31} & \res{20.13}{3.03} & \res{1.75}{1.24} & \res{34.89}{2.89}
& \res{24.37}{5.50} & \res{19.14}{5.11} & \res{12.08}{3.36} & \res{9.19}{1.56} & \res{0.87}{0.65} & \res{30.13}{1.93}
& \res{16.89}{6.30} & \res{15.85}{4.33} & \res{11.53}{4.00} & \res{12.90}{4.45} & \res{0.68}{0.83} & \res{26.84}{9.28}
\\
& {ToG2.0}
& \res{52.45}{1.00} & \res{45.18}{0.84} & \res{30.06}{1.06} & \fir{41.07}{2.14} & \res{77.02}{1.87} & \sed{70.21}{3.01}
& \sed{56.62}{0.91} & \res{53.91}{1.91} & \res{29.36}{2.34} & \sed{31.17}{1.16} & \fir{79.91}{1.93} & \res{57.14}{2.75}
& \res{50.53}{4.72} & \res{52.23}{1.20} & \sed{37.04}{0.52} & \fir{35.07}{1.63} & \sed{74.58}{2.14} & \sed{72.21}{1.87}
\\
& {PoG}
& \res{17.67}{1.68} & \res{17.70}{1.08} & \res{15.27}{0.80} & \res{20.67}{1.37} & \res{2.46}{0.35} & \res{32.77}{3.18}
& \res{24.54}{3.07} & \res{15.31}{4.27} & \res{11.14}{2.27} & \res{10.12}{0.89} & \res{1.39}{0.28} & \res{34.46}{1.66}
& \res{20.23}{2.25} & \res{17.93}{3.51} & \res{18.67}{1.76} & \res{22.17}{1.63} & \res{0.68}{0.83} & \res{34.32}{2.73}
\\
& {R2KG}
& \res{26.75}{1.52} & \res{26.91}{1.08} & \res{20.73}{2.11} & \res{23.22}{2.31} & \res{34.04}{3.82} & \res{25.96}{5.93}
& \res{36.97}{2.14} & \res{26.56}{2.84} & \res{22.28}{1.77} & \res{16.75}{1.54} & \res{30.87}{4.50} & \res{28.75}{2.49}
& \res{31.29}{1.47} & \res{25.80}{1.84} & \res{18.37}{1.96} & \res{19.28}{3.96} & \res{30.51}{6.95} & \res{39.26}{1.13}
\\
& {HippoRAG2.0}
& \res{43.37}{0.00} & \res{55.40}{0.00} & \res{40.85}{0.30} & \res{24.43}{0.33} & \fir{81.05}{0.43} & \res{67.66}{1.59}
& \res{43.36}{0.31} & \res{56.72}{0.38} & \sed{36.51}{0.33} & \res{23.49}{0.38} & \res{73.29}{0.43} & \res{68.21}{0.44}
& \res{43.41}{0.19} & \res{64.87}{0.39} & \res{32.14}{0.00} & \res{32.90}{0.74} & \fir{76.27}{0.00} & \res{68.53}{0.21}
\\
& {EvoReasoner}
& \sed{53.73}{0.59} & \sed{62.16}{0.98} & \fir{49.58}{2.28} & \sed{35.84}{2.90} & \sed{78.42}{2.39} & \fir{77.02}{4.13}
& \res{55.97}{0.67} & \res{60.94}{1.78} & \fir{40.27}{1.12} & \fir{35.06}{1.39} & \res{70.75}{2.52} & \fir{72.86}{2.86}
& \sed{56.29}{0.98} & \sed{66.42}{1.29} & \fir{42.55}{1.63} & \sed{34.35}{1.63} & \res{74.24}{4.21} & \fir{80.63}{0.91}
\\
\bottomrule
\end{tabularx}
}
\end{center}
\end{table*}